\documentclass[twocolumn]{fairmeta}

\usepackage{microtype}
\usepackage{graphicx}
\usepackage{csquotes}
\usepackage{afterpage}
\usepackage{float}
\usepackage{tablefootnote}
\usepackage{subcaption}
\usepackage{longtable}
\usepackage{caption}
\usetikzlibrary{positioning}
\definecolor{gblue9}{rgb}{0.2, 0.5, 0.7}

\usepackage{amsmath}
\usepackage{amssymb}
\usepackage{mathtools}
\usepackage{amsthm}
\usepackage{xspace}
\usepackage{colortbl}
\usepackage{multirow}
\usepackage{enumitem}
\usepackage{wrapfig}
\usepackage{nicefrac}

\makeatletter
\DeclareRobustCommand\onedot{\futurelet\@let@token\@onedot}
\def\@onedot{\ifx\@let@token.\else.\null\fi\xspace}
\makeatother

\newcommand{\cmark}{\checkmark}
\newcommand{\xmark}{\texttimes}

\newcommand{\first}[1]{\textbf{\textcolor{red}{#1}}}
\newcommand{\second}[1]{\textbf{\textcolor{blue}{#1}}}

\newcommand{\Benchmark}{PresentBench}
\newcommand{\BenchmarkSize}{238}

\AtBeginDocument{%
}

\lstdefinelanguage{Markdown}{
    moredelim=[l][\bfseries]{\#},
    moredelim=[l][\bfseries]{\#\#},
    moredelim=[l][\bfseries]{\#\#\#},
    moredelim=[s][\bfseries]{**}{**},
    morecomment=[l]{>},
    sensitive=false,
}

\title{
    \Benchmark: A Fine-Grained Rubric-Based Benchmark for Slide Generation
}

\author[1]{Xin-Sheng Chen}
\author[1]{Jiayu Zhu}
\author[1]{Pei-lin Li}
\author[1]{Hanzheng Wang}
\author[1]{Shuojin Yang 
    \thanks{Corresponding author. \email{yangshuojin@mail.tsinghua.edu.cn}.}
}
\author[1]{Meng-Hao Guo}

\affiliation[1]{Tsinghua University}
\contribution[]{Project Page: \url{https://presentbench.github.io}}

\abstract{

Slides serve as a critical medium for conveying information in presentation-oriented scenarios such as academia, education, and business. 
Despite their importance, creating high-quality slide decks remains time-consuming and cognitively demanding.
Recent advances in generative models, such as Nano Banana Pro, have made automated slide generation increasingly feasible.
However, existing evaluations of slide generation are often coarse-grained and rely on holistic judgments, making it difficult to accurately assess model capabilities or track meaningful advances in the field.
In practice, the lack of fine-grained, verifiable evaluation criteria poses a critical bottleneck for both research and real-world deployment.
In this paper, we propose \Benchmark{}, a fine-grained, rubric-based benchmark for evaluating automated real-world slide generation.
It contains \BenchmarkSize{} evaluation instances, each supplemented with background materials required for slide creation.
Moreover, we manually design an average of 54.1 checklist items per instance, each formulated as a binary question, to enable fine-grained, instance-specific evaluation of the generated slide decks.
Extensive experiments show that \Benchmark{} provides more reliable evaluation results than existing methods, and exhibits significantly stronger alignment with human preferences.
Furthermore, our benchmark reveals that NotebookLM significantly outperforms other slide generation methods, highlighting substantial recent progress in this domain.

}

\begin{document}

\renewcommand{\thefootnote}{\fnsymbol{footnote}}
\setcounter{footnote}{1}  

\maketitle

\renewcommand{\thefootnote}{\arabic{footnote}}
\setcounter{footnote}{0}

\section{Introduction}

\begin{figure}[t]
  \centering
  \includegraphics[width=\columnwidth]{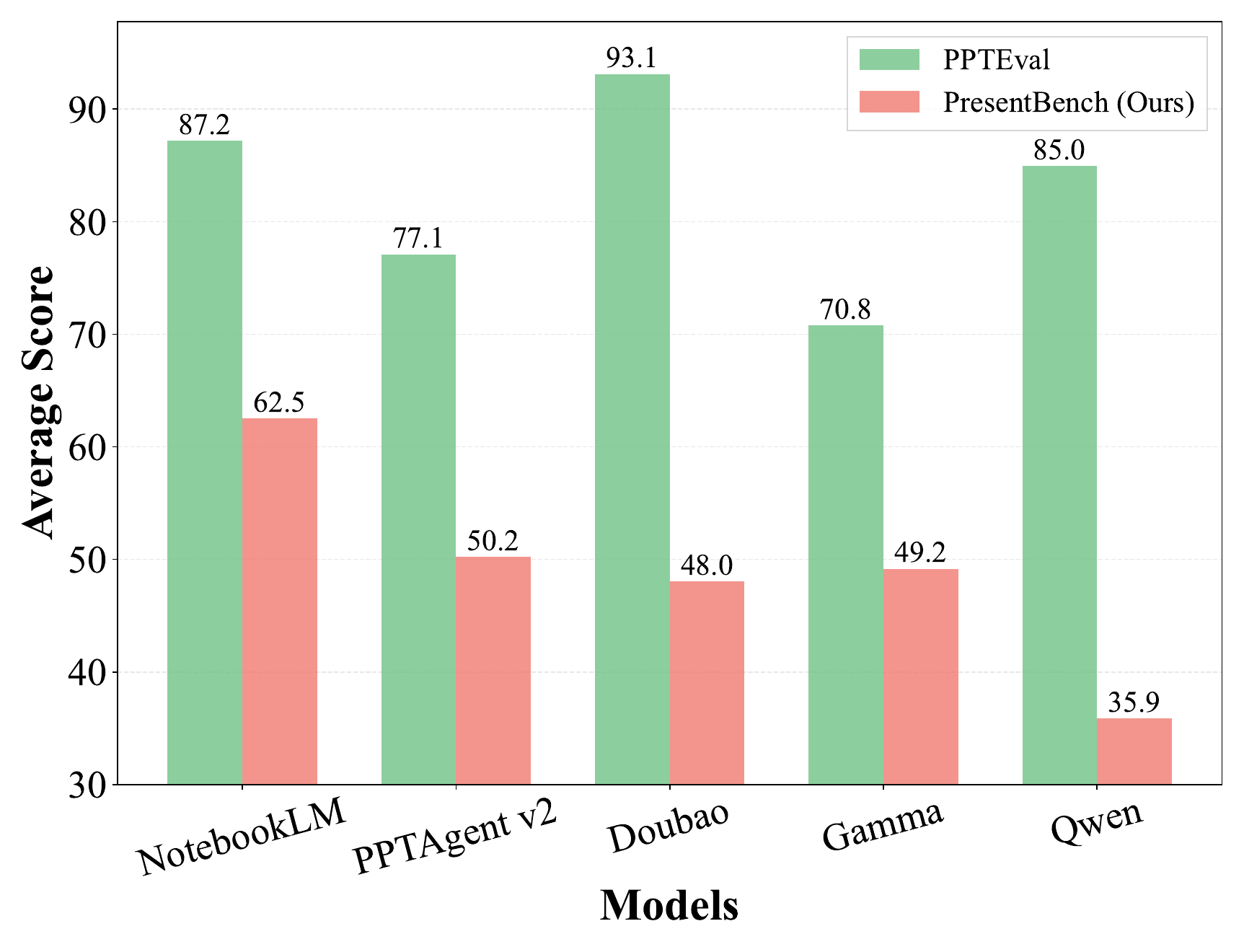} 
  \caption{
    Performance comparison of various slide generation systems on the PPTEval evaluation framework and \Benchmark{}.
    \Benchmark{} adopts a stricter scoring scheme and poses a greater challenge to slide generation systems.
  }
  \label{fig:average_scores_compare_with_ppteval}
\end{figure}

Slides are a core medium for conveying information across domains such as academia, education, and business.
They are essential for knowledge dissemination, decision-making, and professional reporting. 
Despite their widespread use and importance, producing high-quality slide decks remains both time-consuming and cognitively demanding, as it requires careful content selection, coherent structural organization, effective visual design, and strict adherence to background materials.
Meanwhile, intelligent systems are evolving from dialogue-centric chatbots~\cite{DBLP:journals/corr/abs-2303-08774,DBLP:journals/corr/abs-2406-12793,DBLP:journals/corr/abs-2407-21783,DBLP:journals/corr/abs-2407-10671,DBLP:journals/corr/abs-2510-13795} into task-oriented agents~\cite{ DBLP:conf/iclr/YaoZYDSN023,DBLP:conf/iclr/HongZCZCWZWYLZR24,DBLP:conf/iclr/0001LSXTZPSLSTL25} capable of handling complex professional tasks.
With this paradigm shift, slide generation has emerged as one of the most critical and challenging frontiers for AI office assistants. 
However, as slide generation systems continue to evolve, a fundamental question remains largely unresolved:
\textbf{how can we reliably and comprehensively evaluate the quality of automatically generated slide decks?}

\begin{figure*}[t]
    \centering
    \includegraphics[width=1.00\linewidth]{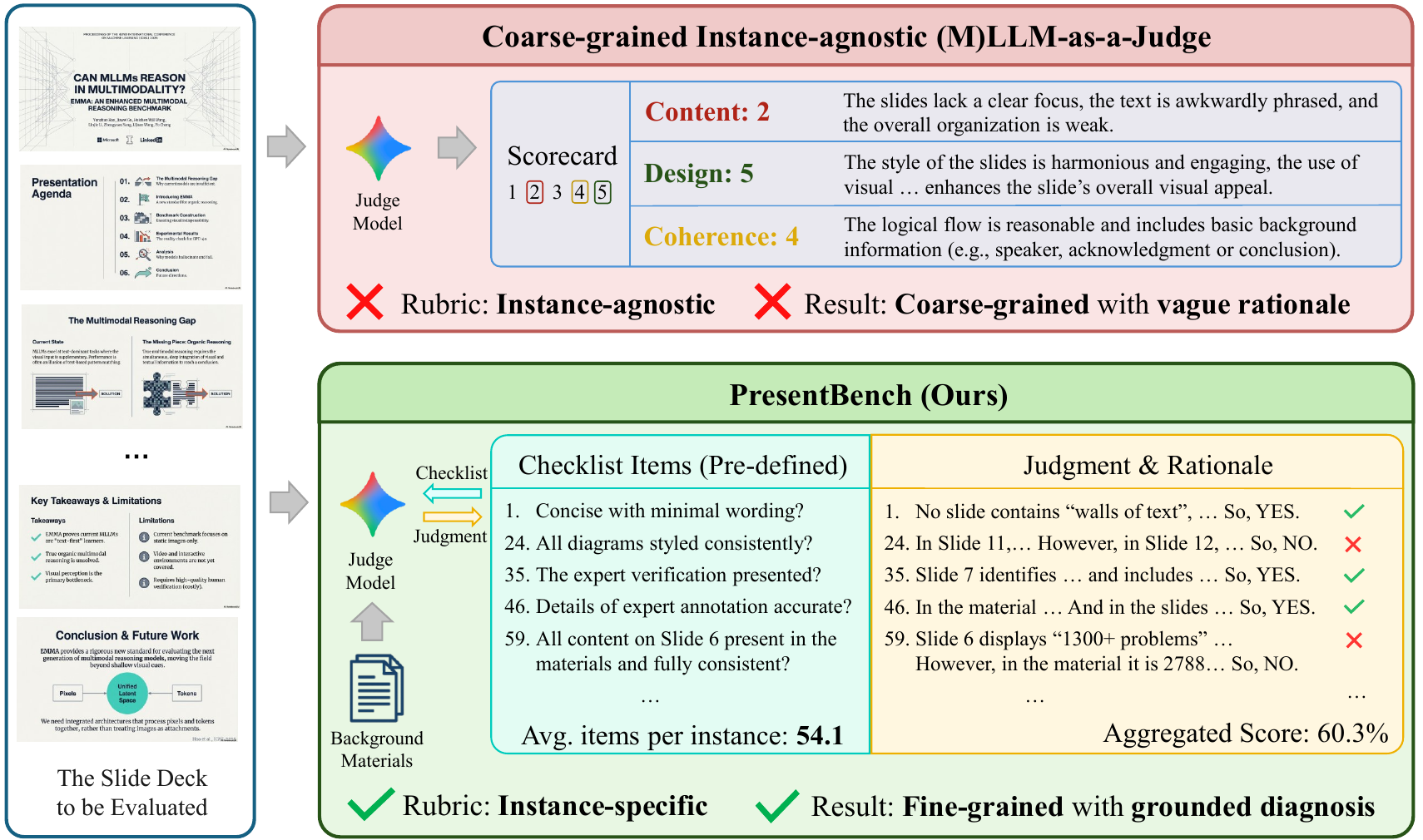}
    \caption{Comparison of coarse-grained, instance-agnostic (M)LLM-as-a-Judge evaluation frameworks and \Benchmark{}.} 
    \label{fig:comparison_our_benchmark_and_coarse_grained}
\end{figure*}

Evaluating slide generation is inherently challenging. Unlike traditional unimodal generation tasks, slide generation is a tightly coupled multimodal problem that involves text, images, and layout simultaneously.
A high-quality slide deck must not only accurately extract and synthesize key information from background materials, but also present the information in a logically coherent structure with appropriate visual design, while strictly preserving factual correctness and quantitative fidelity.
These requirements make it difficult to assess slide quality using static, instance-agnostic metrics or coarse (M)LLM-as-a-judge approaches that merely ask whether a slide deck is well-designed.
As a result, it is essential to establish fine-grained rubrics tailored to each slide deck instance.

As shown in \cref{tab:ppt_benchmark_comparison}, existing evaluation benchmarks for slide generation suffer from several critical limitations. 
First, much prior work focuses on isolated subtasks, such as single-slide generation~\cite{huang2025pptbench} or localized editing~\cite{ofengenden2025pptarena}, rather than evaluating the full end-to-end slide deck authoring process, resulting in a mismatch between evaluation settings and real-world usage.
Second, some benchmarks, such as SlidesBench~\cite{DBLP:conf/cvpr/GeWZPSTSSFND25}, adopt open-ended or reference-free settings without grounding the task in concrete background materials, making it difficult to verify whether slide generation systems truly understand and faithfully follow user-provided inputs.
Third, some evaluation frameworks, such as PPTEval~\cite{DBLP:conf/emnlp/ZhengGKZZZLLHS25}, adopt instance-agnostic scoring schemes, typically relying on the (M)LLM-as-a-Judge paradigm~\cite{DBLP:conf/nips/ZhengC00WZL0LXZ23,DBLP:conf/icml/ChenCZWLZZ00024} that poses the same set of questions for all slide decks, as illustrated in \cref{fig:comparison_our_benchmark_and_coarse_grained}.
As a result, such evaluations fail to account for instance-specific content, including background materials and task requirements, making it difficult to assess whether a model truly follows the intended input and task context.
Fourth, even when explicit evaluation criteria are provided, they are often too coarse-grained to support reliable and informative assessment, as shown in \cref{fig:comparison_our_benchmark_and_coarse_grained}.
Each rubric typically entangles multiple dimensions, including content completeness, factual correctness, and presentation quality, imposing a high cognitive burden on the judge model and leading to subjective judgments with limited verifiability and interpretability.
Furthermore, these aggregate scores obscure specific failure modes, limiting their utility for error diagnosis and model iteration.

\begin{table*}
\centering
\caption{Comparison of slide generation benchmarks.}
\label{tab:ppt_benchmark_comparison}
\resizebox{1.00\textwidth}{!}{%
\begin{tabular}{l|cccc}
  \toprule
    \textbf{Benchmark} & \textbf{Full Deck Generation} & \textbf{Background-based} & \textbf{Instance-specific Criteria} & \textbf{Fine-grained Criteria} \\
  \midrule
    SlidesBench~\cite{DBLP:conf/cvpr/GeWZPSTSSFND25} & \xmark & \xmark & \xmark & \xmark \\
    PPTBench~\cite{huang2025pptbench} & \xmark & \xmark & \xmark & \xmark \\
    PPTEval~\cite{DBLP:conf/emnlp/ZhengGKZZZLLHS25}  & \cmark & \cmark & \xmark & \xmark\\
    PPTArena~\cite{ofengenden2025pptarena}  & \xmark & \xmark & \cmark & \xmark \\
    SlidesGen-Bench~\cite{yang2026slidesgen} & \cmark & \cmark & \cmark & \xmark\\
  \midrule
    \Benchmark{} (Ours) & \cmark & \cmark & \cmark & \cmark\\
  \bottomrule
\end{tabular}%
}
\end{table*}

To address these challenges, we propose \Benchmark{}, a fine-grained, rubric-based, real-world benchmark designed for evaluating automated slide generation.
It consists of \BenchmarkSize{} expert-curated evaluation instances spanning five representative application domains, each paired with authentic background materials required for slide creation.
For each instance, we design task-specific generation instructions with explicit constraints covering objectives, structure, content fidelity, visual layout, quantitative accuracy, and audience definition.
Crucially, \Benchmark{} introduces a fine-grained checklist-based evaluation framework.
On average, each instance is associated with more than 50 specifically designed atomic evaluation items, 
covering five complementary aspects:
(1) presentation fundamentals, which assess logical flow, conciseness, linguistic quality, and appropriateness for the target scenario;
(2) visual design and layout, which evaluate visual quality, legibility, and layout appropriateness;
(3) content completeness, which measures whether all required content specified in the instruction is included;
(4) content correctness, which verifies the correctness of the required content;
and (5) content fidelity, which enforces strict grounding in the background materials to prevent hallucinated or fabricated content.   
By decomposing slide evaluation into verifiable, instance-specific checklist items and aggregating the resulting decisions via principled scoring mechanisms, \Benchmark{} provides reliable, interpretable, and human-aligned evaluation signals.
As previewed in \cref{fig:average_scores_compare_with_ppteval}, coarse-grained, instance-agnostic evaluation frameworks such as PPTEval tend to produce overly optimistic scores. 
By contrast, \Benchmark{} adopts a stricter, more fine-grained rubric that poses a greater challenge to slide generation systems and better reveals their failure modes. 

Experiments further demonstrate that \Benchmark{} yields more accurate evaluations than existing methods and exhibits substantially stronger agreement with human preferences. 
Our results reveal clear performance differences among current slide generation systems, with NotebookLM~\cite{notebooklm} consistently outperforming others across scenarios, highlighting its substantial progress in this field.

In summary, we make the following contributions:

\begin{itemize}
    \item 
        We introduce \Benchmark{}, a fine-grained rubric-based benchmark for evaluating automated slide generation in real-world presentation scenarios. \Benchmark{} consists of \BenchmarkSize{} expert-curated instances, and for each instance we design over 50 atomic checklist items specifically tailored to the task, enabling instance-specific, verifiable, and fine-grained assessment of slide quality.
    
    \item  
        Experiments across a wide range of slide generation methods demonstrate that \Benchmark{} yields evaluation results that align significantly more closely with human preferences than existing evaluation frameworks. Moreover, it clearly quantifies differences in model performance and reveals both recent progress, most notably by NotebookLM, and the remaining limitations of current approaches.

\end{itemize}

\section{Related Work}

\subsection{Slide Generation Benchmarks}

Current evaluations of slide generation primarily adopt the LLM-as-a-Judge paradigm~\cite{DBLP:journals/corr/abs-2412-05579}, where large language models directly provide holistic scores for generated outputs without requiring a single canonical reference.
For example, PPTEval~\cite{DBLP:conf/emnlp/ZhengGKZZZLLHS25} employs multimodal large language models to assess slide decks along dimensions such as content quality, visual design, and coherence, 
while SlidesBench~\cite{DBLP:conf/cvpr/GeWZPSTSSFND25} complements reference-based evaluation with reference-free assessment to measure intrinsic design quality. 
To better align automated evaluation with human judgment, recent works~\cite{DBLP:conf/iclr/LiSYF0024,DBLP:conf/emnlp/KimSLLSWNL0S24} develop judge models trained to match human or strong-judge preferences while supporting flexible evaluation criteria.
However, although human alignment can partially mitigate model subjectivity, the evaluation process still imposes a substantial cognitive burden on the judge model, thereby limiting evaluation accuracy.

\subsection{Rubric-Based Evaluation}

Simple holistic evaluation metrics often fail to comprehensively characterize model performance. 
To address this limitation, GenExam~\cite{DBLP:journals/corr/abs-2509-14232} adopts an exam-style evaluation paradigm, designing explicit scoring points to verify semantic correctness and visual plausibility of cross-disciplinary text-to-image generation results on an item-by-item basis; 
MMMG~\cite{luo2025mmmg} targets knowledge image generation by introducing explicit knowledge graphs as intermediate representations and assessing semantic fidelity in multi-entity, multi-relation scenarios using graph alignment; 
OneIG-Bench~\cite{chang2025oneigbench} systematically decomposes image generation capabilities along dimensions such as semantic alignment, text rendering, knowledge and reasoning, style, and diversity, providing a comprehensive multi-dimensional diagnostic view.
However, for slide generation tasks, there is still a lack of a systematic evaluation standard with similarly strong fine-grained diagnostic capability, which significantly limits in-depth analysis and reliable comparison of model performance.

\section{\Benchmark{}}

\begin{figure*}[t]
    \centering
    \includegraphics[width=\linewidth]{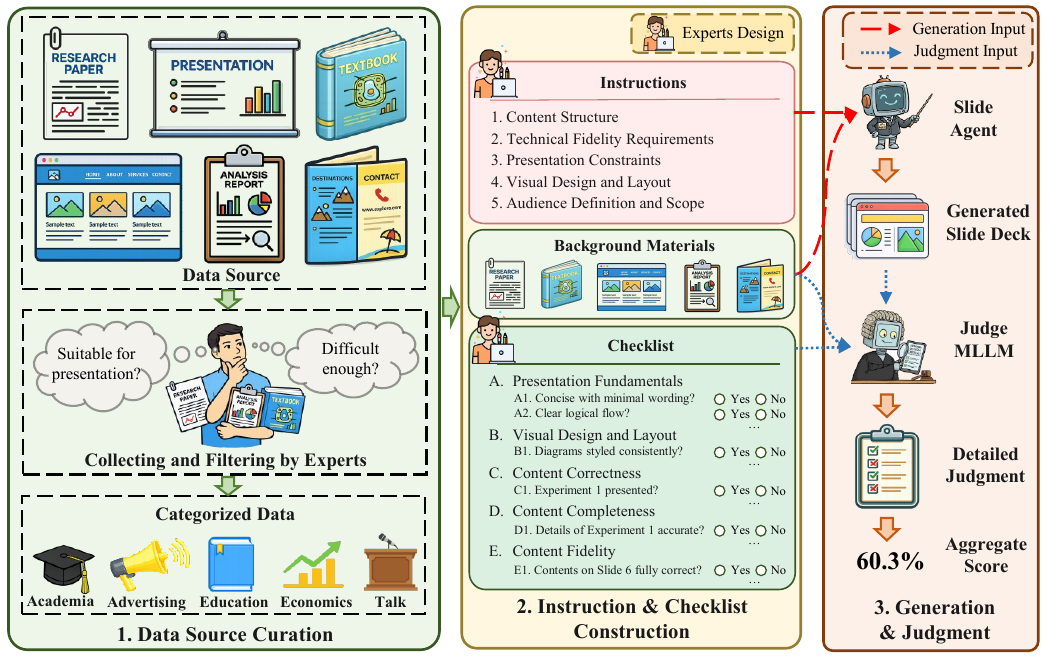}
    \caption{
        The construction and evaluation workflow of \Benchmark{}.
        (a) Benchmark construction (left and middle): Experts collect and filter diverse data sources into five domains (e.g., Academia and Education), and design instance-specific, fine-grained instructions and checklists to formulate concrete slide generation tasks.
        (b) Evaluation (right): The slide agent generates a slide deck based on the instructions and corresponding materials, followed by an automated assessment in which a judge MLLM utilizes the structured checklist to conduct the evaluation and yields a report with per-item decisions and supporting evidence, which is then aggregated into an overall score.
    } 
    \label{fig:pipeline_of_benchmark}
\end{figure*}

In this section, we introduce \Benchmark, a benchmark designed for evaluating slide generation capabilities. 
Through a series of meticulously designed fine-grained generation instructions, this benchmark guides slide generation systems through the complete synthesis process, from initial content comprehension to final layout presentation. 
In conjunction with a corresponding set of evaluation rubrics, the benchmark performs a systematic and comprehensive assessment of the generated slide decks.
\cref{fig:pipeline_of_benchmark} summarizes the overall construction and evaluation workflow of \Benchmark.

\subsection{Data Source Curation}
Our dataset is constructed from a diverse collection of real-world slide-related materials spanning a wide range of domains and presentation contexts. The dataset comprises \BenchmarkSize{} high-quality evaluation instances, covering five thematic categories: 
Academia, Education, Economics, Talk, and Advertising.
This topical diversity ensures broad coverage of both professional and everyday office use cases.

For each instance, we collect and curate background materials from authoritative and publicly available sources, including top-tier conference papers (e.g., those from ICLR, ICML, CVPR, USENIX), university course textbooks (e.g., those from CMU, MIT), financial and economic reports from major corporations and institutions (e.g., Microsoft, JPMorgan, OECD, World Bank), as well as public speeches, educational talks, and commercial brochures. 
Depending on the source, the background materials may comprise one or more documents.
Notably, during the data collection process, we prioritize sources that originally contained paired background materials and slides; although we utilize only the former, this selection strategy ensures that every test case represents a genuine, real-world scenario necessitating slide generation.

All data sources are manually inspected and filtered by experts to ensure correctness, relevance, and suitability for slide generation. 
In addition, we assess the difficulty of the corresponding slide-authoring scenarios and exclude overly trivial instances (e.g., generating slides from a very short speech transcript).
We further verify data accessibility, language consistency (primarily English with a small portion of Chinese samples), and conduct best-effort checks of usage terms to reduce potential copyright risks.
This rigorous curation process guarantees that \Benchmark{} represents realistic, legally mindful, and challenging end-to-end slide generation tasks grounded in authentic materials.

\subsection{Instructions}

To rigorously evaluate end-to-end slide deck generation across diverse scenarios, we design a highly constrained instruction tailored to each evaluation instance.
The instruction defines the task as generating a slide deck for a specified scenario, requiring all content to be grounded in the provided background materials and prohibiting the inclusion of unsupported information. 
An example instruction is shown in \cref{sec:instruction_example}.

\paragraph{Structural and content constraints.}
The instruction specifies strict structural requirements, including a slide-count range (e.g., 16--20 slides) and an ordered, instance-specific set of mandatory sections (e.g., title, agenda, background, methodology, experimental results, discussion, and conclusion). 
For each section, the instruction further specifies the expected content scope and key points to cover, ensuring sufficient technical depth and a coherent narrative flow.
This design constrains slide generation systems to follow realistic presentation structures rather than producing loosely organized or superficial content. 
During evaluation, the completeness and correctness of these requirements are checked item-by-item using our fine-grained, instance-specific checklist (\cref{sec:checklist}).

\paragraph{Faithfulness and accuracy guarantees.}
To ensure faithfulness to the background materials, the instruction requires that all content be derived exclusively from the provided inputs and prohibits fabrication or alteration of reported information. 
Each slide deck must include sufficient quantitative details, and all quantitative content must be traceable to specific locations in the background materials and exactly match the original values unless explicitly marked as conceptual illustrations.

\paragraph{Presentation quality and brevity constraints.}
The instruction specifies language and presentation style (concise, bullet-point-based), requires coherent and on-topic content, and excludes non-slide artifacts (e.g., scripts or design rationales), aligning the task with real-world slide authoring practices.

\begin{figure}[!t]
  \centering
  \includegraphics[width=\columnwidth]{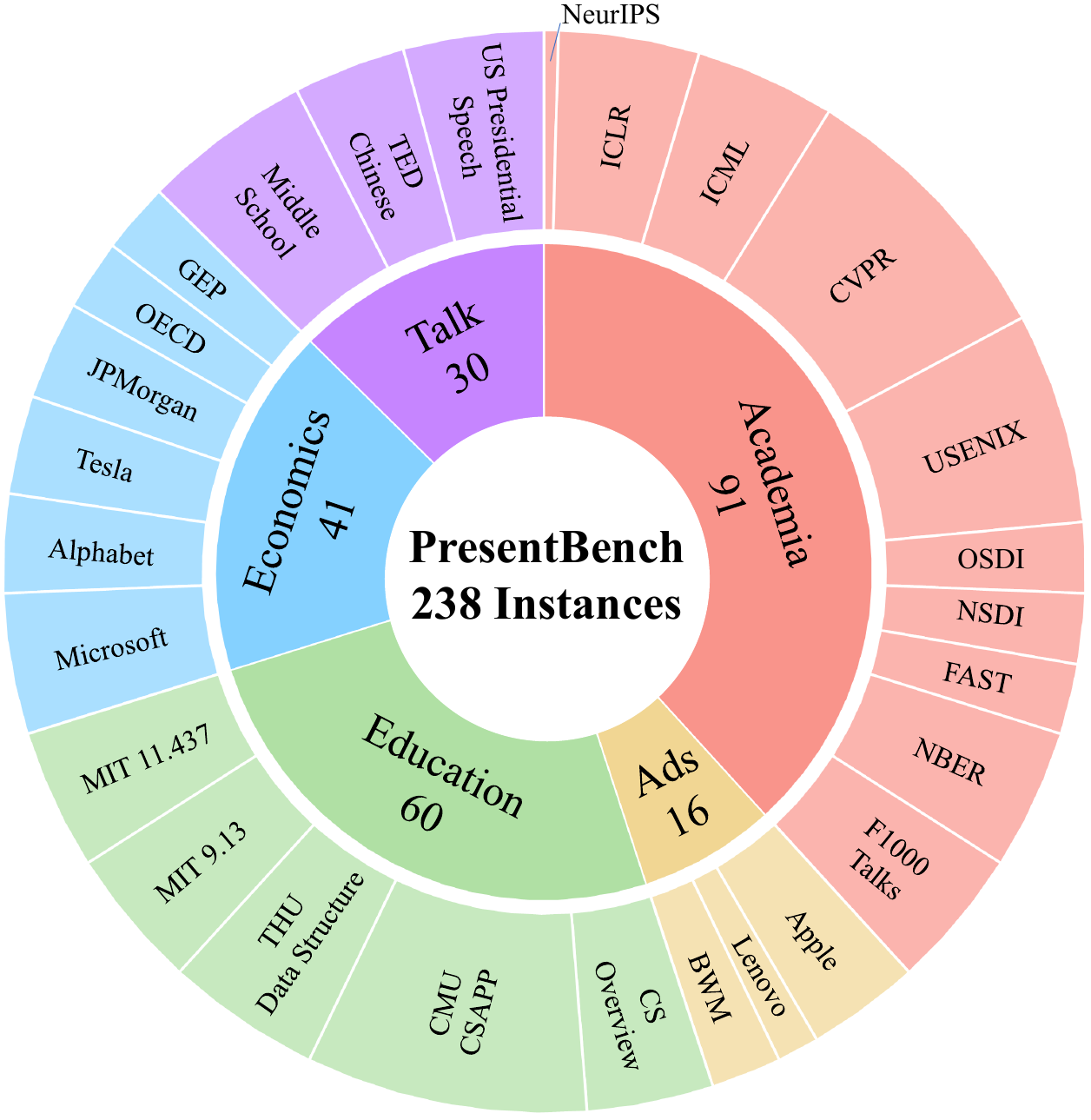} 
  \caption{
    The distribution of \Benchmark{}. ``Ads'' denotes the Advertising category.
  }
  \label{fig:pie_task}
\end{figure}

\paragraph{Visual and layout requirements.}
To assess cross-modal organization and visual planning abilities, the instruction specifies detailed design requirements, including consistent visual style, legible typography, balanced text-visual layouts, and appropriate use of high-quality images, charts, and diagrams. 
All visualized content must be clearly annotated (axes, units, legends) and traceable to specific figures, tables, or sections in the background materials.

\paragraph{Audience and tone specification.}
The instruction also specifies the target audience and an appropriate tone, ensuring alignment with the intended scenario.

Taken together, these comprehensive, instance-specific constraints transform slide generation into a structured and verifiable task that mirrors real-world authoring workflows. 
They also allow us to assess slide generation systems against verifiable criteria, thereby enabling reliable, diagnostic evaluations.

\subsection{Fine-Grained Evaluation Checklist} \label{sec:checklist}

To enable fine-grained, verifiable, and interpretable evaluation, we design a comprehensive checklist-based scoring framework, which decomposes slide quality assessment into a set of atomic, item-level criteria. 
The checklist is organized into two complementary tiers, namely material-independent and material-dependent tiers, allowing us to jointly assess intrinsic slide quality and fidelity to the provided background materials.
An example of the checklist is provided in \cref{sec:checklist_example}.

\begin{figure}[!t]
    \centering
    \includegraphics[scale=0.33]{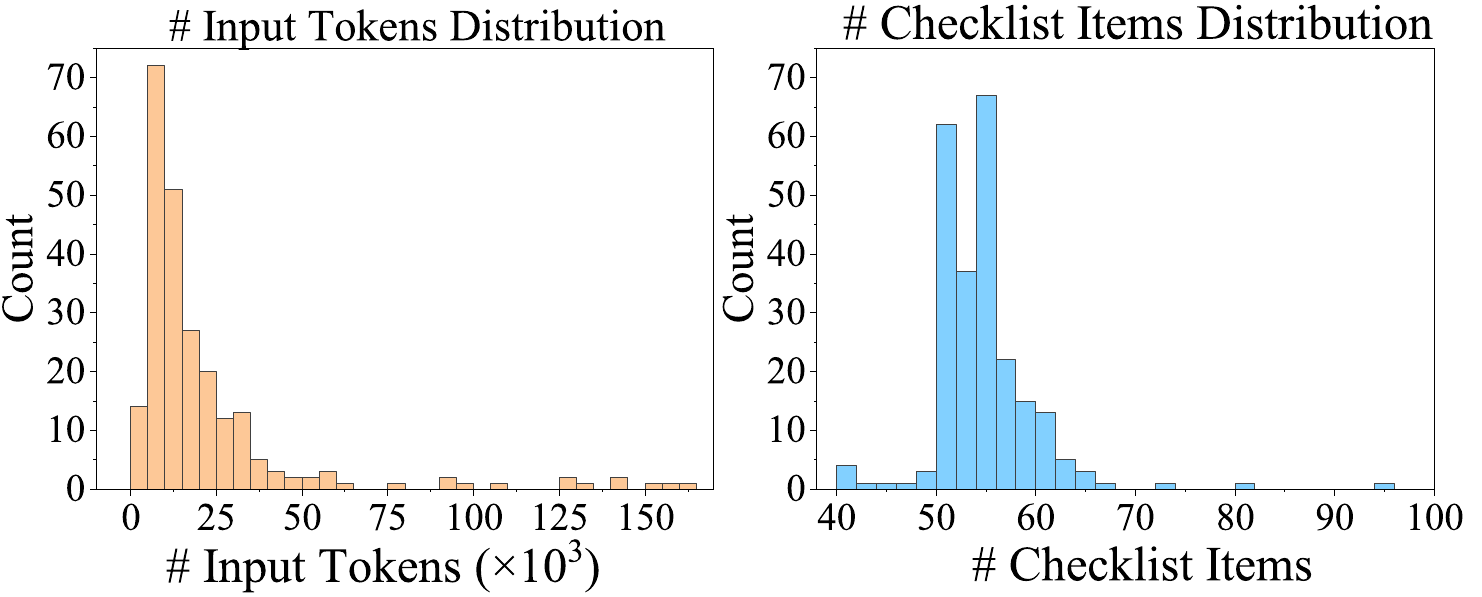}
    \caption{Distribution of the number of input tokens (left) and the number of checklist items (right).} 
    \label{fig:statistics_of_input_and_checklist}
\end{figure}

\subsubsection{Material-Independent Checklist}

The material-independent checklist evaluates the intrinsic quality of the generated slide deck without requiring access to the original background material. It consists of two dimensions:

\begin{enumerate}
    \item \textbf{Presentation Fundamentals.}  
    This dimension assesses whether the slide deck presents information in a clear, coherent, and appropriate manner. 
    It examines properties such as theme clarity, cross-slide logical flow, title-content relevance, conciseness, and safety (no harmful or biased content).

    \item \textbf{Visual Design and Layout.}  
    This dimension evaluates visual consistency and layout quality, including design uniformity (fonts, colors, and layout), balance between text and visuals, relevance and moderation of decorative elements, layout appropriateness (e.g., avoiding empty or title-only slides), and the absence of visual clutter. It further checks for text or visual element overlap, image quality, appropriate use of charts or diagrams, visual appeal, bullet-point density constraints, and font size legibility.
\end{enumerate}

\begin{table}[t]
\centering
\caption{
    Key statistics of \Benchmark{}.
    ``Fundamentals'', ``Visual'', ``Completeness'', ``Correctness'', and ``Fidelity'' 
    denote 
    ``Presentation Fundamentals'', 
    ``Visual Design and Layout'',
    ``Content Completeness'',
    ``Content Correctness'', 
    and ``Content Fidelity'', 
    respectively.
    }
\label{tab:statistics}
\begin{tabular}{lc}
  \toprule
    Statistics                   & Number  \\
  \midrule
    Total evaluation instances     & \BenchmarkSize{} \\
  \midrule
    Academia evaluation instances  & 91 \\
    Advertising evaluation instances  & 16 \\
    Economics evaluation instances  & 41 \\
    Education evaluation instances  & 60 \\
    Talk evaluation instances     & 30 \\
  \midrule
    English evaluation instances   & 219 \\
    Chinese evaluation instances   & 19 \\
  \midrule
    Average input tokens           & $22.2 \times 10^3$ \\
     - Average instruction tokens & $2.3 \times 10^3$  \\
     - Average material tokens    & $19.9 \times 10^3$ \\
    Average material pages \tablefootnote{Non-PDF materials are excluded from page counts.}        & 34.0  \\
  \midrule
    Average checklist items     & 54.1 \\
     - Average Fundamentals items         & 13.0    \\
     - Average Visual items         & 17.0    \\
     - Average Completeness items         & 12.6    \\
     - Average Correctness items         & 11.5    \\
     - Average Fidelity items         & Dynamic\tablefootnote{The number of Fidelity items varies across slide decks; see \cref{sec:dependent_checklist} for details.}    \\
  \bottomrule
\end{tabular}
\end{table}

\subsubsection{Material-Dependent Checklist} \label{sec:dependent_checklist}

The material-dependent checklist evaluates whether the generated slide deck faithfully follows and accurately reflects the provided background materials. It consists of three dimensions:

\begin{enumerate}
    \item \textbf{Content Completeness.} 
    This dimension verifies coverage using an instance-specific, instruction-derived checklist, examining whether each required section appears in the intended order. 
    It further assesses whether the key points mandated for each section are sufficiently addressed, ensuring no instruction-specified content is omitted.

    \item \textbf{Content Correctness.}
    This dimension evaluates whether the instruction-mandated content is presented correctly, i.e., consistent with the provided background materials. 
    Importantly, each required content item must be both \emph{present} and \emph{correct}: if a mandated item is missing from the slide deck, it is also counted as an error under this dimension.    

    \item \textbf{Content Fidelity.}  
    This dimension verifies strict adherence to the provided background materials by checking the slides on a page-by-page basis.
    All presented information must be traceable to, and consistent with, the sources. Any unsupported, contradictory, or newly introduced detail is treated as hallucinated and counted as an error.

    Content Fidelity is complementary to Content Correctness: if slide generation is viewed as a retrieval task, Content Correctness is analogous to recall, whereas Content Fidelity is analogous to precision.
    
    We implement this check with page-level checklist items (one per slide page), so the number of items varies across slide decks.
    For clarity, the checklist-size statistics reported exclude this component.
    
\end{enumerate}

\subsubsection{Scoring and Aggregation Protocol} \label{sec:evaluation_metrics}

During evaluation, each checklist item is verified independently, and a binary verdict (e.g., satisfied or violated) is assigned along with localized evidence, such as the relevant slide indices and affected content elements.

We evaluate the overall performance of a slide deck by computing an aggregate score. Specifically, for each evaluation dimension $i$ where $1 \leq i \leq 5$, we first calculate the normalized dimension score $s_i$:
\begin{equation}
    \label{eq:evaluation_dimension}
    s_i = \frac{1}{N_i} \sum_{j=1}^{N_i} s_{i,j},
\end{equation}
where $N_i$ is the number of checklist items in dimension $i$, and $s_{i,j} \in \{0,1\}$ is a binary indicator of whether the $j$-th item is satisfied.

The final score for each evaluation instance is computed as the average of the five dimension scores:
\begin{equation}
    s = \frac{1}{5} \sum_{i=1}^{5} s_i .
\end{equation}

\subsection{Overview of \Benchmark{}}

The main statistics of \Benchmark{} are presented in \cref{tab:statistics}. 
The dataset covers five representative real-world slide-creation domains and is predominantly in English, with a small Chinese subset, reflecting both practical usage and multilingual evaluation needs.
\cref{fig:pie_task} shows the domain distribution. More detailed breakdowns of the dataset composition are provided in \cref{tab:detailed_composition_of_the_dataset}.

The average input length is $22.2 \times 10^3$ tokens, with an average of 34.0 material pages.
This scale implies that slide generation systems must handle long-context materials and distill them into a coherent slide deck, rather than relying on short prompts or surface-level paraphrasing, thereby making the task challenging in practice.
Meanwhile, each instance is associated with a fine-grained checklist containing 54.1 binary items on average, covering five dimensions, which together support comprehensive, diagnostic, and verifiable assessment.
The distributions of input tokens and checklist sizes are shown in \cref{fig:statistics_of_input_and_checklist}.

\section{Experiments}

\subsection{Experimental Setup}

We evaluate several widely used slide generation products currently available on the market, including NotebookLM~\cite{notebooklm}, Manus 1.6~\cite{manus}, Gamma~\cite{gamma}, Doubao~\cite{doubao}, Tiangong~\cite{tiangong}, Zhipu~\cite{zhipu}, and Qwen~\cite{qwen}, as well as an open-source slide agent framework, PPTAgent~\cite{DBLP:conf/emnlp/ZhengGKZZZLLHS25}. 

For each method, we provide the same background materials together with the corresponding instructions, and require the model to produce a complete slide deck (either in PDF or PPTX format). 
To control for the effects of additional user interactions, we uniformly adopt default settings for all optional operations within each framework, such as template selection and layout configuration.
For PPTAgent, we use version v2.0.0 and generate slides in its ``freeform mode''. We employ doubao-seed-1-8-251228 as the ``research agent'' and the ``design agent'', and gemini-3-pro-image-preview as the image generation model.

We use gemini-3-flash-preview as the judge model in all evaluations. 
For slide decks produced in PPTX format, we convert them to PDF before providing them to the judge model.
To reduce the cognitive burden on the judge model and thereby improve evaluation reliability, we evaluate each checklist item in a separate call.
Because effective information compression is necessary for high-quality slides, we truncate any output exceeding the instruction-specified page limit to that maximum during evaluation.

We follow the metric definition in \cref{sec:evaluation_metrics} and report the mean aggregated score over \BenchmarkSize{} instances.

\subsection{Main Results}

\begin{table*}[t]
\centering
\setlength{\tabcolsep}{15pt}
\caption{Comparative results across five domains. The highest scores are highlighted in \first{red}, and the second-highest scores are highlighted in \second{blue}.}
\label{tab:domain_results}
\begin{tabular}{l|c|ccccc}
  \toprule
    Method
    & Total
    & Academia
    & Advertising
    & Education
    & Economics
    & Talk \\
  \midrule
    NotebookLM  & \first{62.5} & \first{68.6}  & \first{54.9}  & \first{55.0}  & \first{58.2} & \first{69.2} \\
    Manus 1.6   & \second{57.8} & \second{64.0} & \second{52.4} & 50.7 & \second{52.8} & \second{63.0} \\
    Gamma        & 49.2 & 54.4 & 46.7 & 47.8          & 35.1 & 56.3 \\
    Doubao       & 48.0 & 50.3 & 42.9 & 45.4          & 44.0 & 54.7 \\
    Tiangong     & 54.7 & 59.2 & 44.5 & \second{53.7} & 46.5 & 59.8 \\
    Zhipu        & 53.6 & 57.5 & 41.0 & 52.5          & 47.6 & 59.0 \\
    Qwen         & 35.9 & 39.4 & 31.9 & 36.6          & 26.5 & 38.6 \\
    PPTAgent v2  & 50.2 & 53.3 & 46.7 & 46.1          & 46.1 & 56.6 \\
  \bottomrule
\end{tabular}
\end{table*}

As shown in \cref{tab:domain_results}, across \BenchmarkSize{} real-world slide deck generation instances, \Benchmark{} clearly differentiates current systems and reveals substantial room for improvement in grounded, end-to-end slide authoring.
Overall, NotebookLM achieves the best performance (62.5), followed by Manus 1.6 (57.8), while other strong commercial systems cluster in the mid-range (48--55).
Below we summarize several key observations and findings from these results.

\paragraph{Slide generation remains challenging for current AI systems.}
Even the best-performing system only reaches 62.5 overall, indicating that grounded, end-to-end slide authoring remains far from solved in real use cases. 
One key reason is the long-context, multi-source nature of our tasks: the input averages 22.2k tokens, with 19.9k tokens coming from background materials (about 34 pages on average), which requires slide generation systems to reliably read, select, synthesize, and organize information across many dispersed facts.
In practice, slide generation systems often exhibit characteristic failure modes, including omissions of instruction-mandated points, introducing unsupported claims, misrepresenting key facts or numbers, and losing cross-slide consistency in terminology or emphasis.
We provide concrete illustrations in \cref{sec:case_study}.

\paragraph{Open-source systems lag behind closed-source counterparts.}
Compared to closed-source products, the open-source ecosystem for slide generation remains limited at present.
Moreover, many available open-source systems focus on partial capabilities, such as localized editing or generation, or template filling. In addition, many of them do not provide reliable material-grounded generation.
As a result, our experiments include PPTAgent as a representative open-source system. 
Although PPTAgent is equipped with a leading image generation model (gemini-3-pro-image-preview), it still lags behind the strongest closed-source systems such as NotebookLM and Manus (50.2 vs. 62.5 and 57.8, respectively).
This gap suggests that state-of-the-art performance often depends not only on the backbone model, but also on proprietary end-to-end pipelines, such as slide-specific long-context planning and grounding, layout and design engines, and richer rendering components. Overall, the results indicate a persistent capability gap between open-source and closed-source slide generation systems, highlighting substantial headroom for the open-source community.

\begin{table*}[t]
\centering
\caption{
    Comparative results across five evaluation dimensions. The highest scores are highlighted in \first{red}, and the second-highest scores are highlighted in \second{blue}.
    ``Fundamentals'' and ``Visual \& Layout'' denote ``Presentation Fundamentals'' and ``Visual Design and Layout'', respectively.
}
\label{tab:dimensions_results}
\resizebox{\linewidth}{!}{
\begin{tabular}{l|ccccc}
  \toprule
    Method
    & Fundamentals 
    & Visual \& Layout
    & Content Completeness 
    & Content Correctness 
    & Content Fidelity \\  
  \midrule
    NotebookLM & \first{81.0}  & \first{62.8}  & \second{67.8} & \first{56.0}  & 45.1 \\
    Manus 1.6  & \second{80.1} & \second{53.7} & 63.6          & 46.2          & \second{45.4} \\
    Gamma      & 66.9          & 22.6          & 54.3          & \second{47.7} & \first{54.1} \\
    Doubao     & 71.8          & 40.7          & 58.2          & 34.7          & 34.8 \\
    Tiangong   & 77.7          & 47.2          & \first{68.8}  & 45.7          & 34.3 \\
    Zhipu      & 73.3          & 40.6          & 63.0          & 47.1          & 44.1 \\
    Qwen       & 53.1          & 21.9          & 29.7          & 29.9          & 44.6 \\
    PPTAgent v2 & 79.8          & 44.4          & 60.2          & 37.9          & 28.8 \\
  \bottomrule
\end{tabular}
}
\end{table*}

\paragraph{Visual Design and Layout is a primary bottleneck and differentiator.}
Across methods, Visual Design and Layout scores are low overall, making it a primary bottleneck for slide generation. 
In ~\cref{tab:dimensions_results}, even the best system, NotebookLM, reaches only 62.8, while most methods are in the 40s or below.
By contrast, Fundamentals scores are much higher (many systems cluster around 70--80), suggesting that slide generation systems can often form a reasonable structure but struggle to render it into polished, presentation-ready layouts. 

Meanwhile, Visual Design and Layout also shows the largest cross-model gap, clearly separating top-tier closed systems from the rest. 
Manus ranks second at 53.7, yet it still lags substantially on layout, even though its content-related scores trail NotebookLM by a smaller margin, indicating that strong content generation does not necessarily translate into strong design capability.
Overall, closing this gap will likely require dedicated visual design pipelines and rendering components, not just stronger backbone models.

\paragraph{Material grounding remains challenging.}
As shown in \cref{tab:dimensions_results}, Content Completeness is often notably higher than Content Correctness (e.g., NotebookLM 67.8 vs. 56.0, PPTAgent 60.2 vs. 37.9), suggesting that many methods can produce the right structure and topics but frequently make factual or numerical mistakes when filling in specifics. 
Content Fidelity also remains challenging even for strong systems (e.g., NotebookLM 45.1, Manus 45.4), pointing to persistent ungrounded details and hallucinations.
Overall, these three content-related dimensions jointly indicate that, for current slide generation systems, reliably achieving both broad coverage and factual accuracy in material-grounded slide authoring remains a critical unresolved challenge, even though it is essential for practical usability.

\begin{table*}[t]
\centering
\caption{
    Dimension-wise evaluation results and human alignment.
    ``Fundamentals'', ``Visual \& Layout'', ``Completeness'', ``Correctness'', and ``Fidelity'' denote 
    ``Presentation Fundamentals'', ``Visual Design and Layout'', ``Content Completeness'', ``Content Correctness'', and ``Content Fidelity'', respectively.
}
\label{tab:dimension_spearman}
\resizebox{1.0\textwidth}{!}{
\begin{tabular}{l|ccccc|c}
  \toprule
    Metric Variant &
    Fundamentals &
    Visual \& Layout &
    Completeness &
    Correctness &
    Fidelity &
    Spearman's $\rho$ \\
  \midrule
    Full  & \cmark & \cmark & \cmark & \cmark & \cmark & 0.532 \\
    Material-Independent Only & \xmark    & \xmark    & \cmark & \cmark & \cmark & 0.319 \\
    Material-Dependent Only & \cmark & \cmark & \xmark    & \xmark    & \xmark    & 0.629 \\
    w/o Fundamentals & \xmark    & \cmark & \cmark & \cmark & \cmark & 0.490 \\
    w/o Visual \& Layout  & \cmark & \xmark    & \cmark & \cmark & \cmark & 0.453 \\
    w/o Completeness & \cmark & \cmark & \xmark    & \cmark & \cmark & 0.477 \\
    w/o Correctness  & \cmark & \cmark & \cmark & \xmark    & \cmark & 0.554 \\
    w/o Fidelity & \cmark & \cmark & \cmark & \cmark & \xmark    & 0.673 \\
  \midrule
    Human & -- & -- & -- & -- & -- & 0.664 \\
  \bottomrule
\end{tabular}
}
\end{table*}

\subsection{Human Alignment Results}

We conduct a user study to assess the alignment between our evaluation methods and human preferences. Specifically, approximately 10\% of the full dataset is uniformly sampled across content categories, resulting in 24 samples covering five domains. 
For each sample, human annotators rank the outputs generated by five systems (NotebookLM, Gamma, Doubao, Qwen, and PPTAgent), based on overall quality, resulting in human preference rankings.
We then compare the model rankings produced by different automatic evaluation methods with the human rankings and compute the Spearman's rank correlation coefficient. The final score is reported as the average Spearman correlation across all samples, where a higher value indicates stronger alignment with human preferences.

As shown in \cref{tab:human_alignment_results}, \Benchmark{} achieves a Spearman correlation of 0.532. This significantly outperforms PPTEval~\cite{DBLP:conf/emnlp/ZhengGKZZZLLHS25} (0.303), which adopts an MLLM-as-a-Judge paradigm to directly assign integer scores on a scale of 1 to 5 to the generated slides. 
We also include an MLLM-as-a-Judge Ranking baseline (0.258), where the MLLM is given the same five slide decks as the user and asked to rank them by overall quality.
The higher correlation of \Benchmark{} suggests that its fine-grained checklist mechanism is more consistent with human preferences than the other holistic scoring approaches. The human agreement score (Human, 0.664) serves as an upper bound reference for consistency in this task. 
Overall, \Benchmark{} effectively narrows the gap between automatic and human evaluations, demonstrating its ability to accurately capture human judgments of slide generation quality.

\begin{table}[t]
\centering
\setlength{\tabcolsep}{15pt}
\caption{Human alignment results across different evaluation methods.}
\label{tab:human_alignment_results}
\begin{tabular}{lc}
  \toprule
    Evaluation Method         & Spearman's $\rho$  \\
  \midrule
    \Benchmark{} (Ours)       & 0.532 \\
    PPTEval                   & 0.303 \\
    MLLM-as-a-Judge Ranking   & 0.258 \\
  \midrule
    Human                     & 0.664 \\
  \bottomrule
\end{tabular}
\end{table}

\subsection{Comparison with Other Evaluation Frameworks}

We evaluate the same benchmark dataset and the same generated slide decks using both our fine-grained, rubric-based scoring and PPTEval, rescaling PPTEval’s 1--5 ratings to a 0--100 scale for comparison. 
As shown in \cref{fig:average_scores_compare_with_ppteval}, PPTEval consistently yields higher scores than \Benchmark.

This gap mainly stems from differences in granularity and verifiability. 
PPTEval makes a single, global judgment based on generic criteria, which often misses subtle errors and offers limited diagnostic value. 
\Benchmark{} instead breaks assessment into instance-specific, source-grounded binary checklist items and aggregates them with principled scoring, making errors easier to localize (e.g., missing content, factual issues, or layout and visual deficiencies). 
Together with our human alignment results (\cref{tab:human_alignment_results}), \cref{fig:average_scores_compare_with_ppteval} suggests \Benchmark{} provides a more challenging, diagnostic, and human-aligned measure of material-based slide deck generation.

\subsection{Ablation Study}

\cref{tab:dimension_spearman} reports ablations on different combinations of scoring dimensions. 
Overall, adding complementary dimensions generally improves alignment with human preferences, as reflected by higher Spearman correlations, suggesting that human judgments capture both presentation quality and coarse content signals.

Interestingly, dimensions targeting fine-grained content fidelity and material-grounded correctness do not always yield further gains in human alignment. Rather than undermining their value, this pattern likely reflects a limitation of the user study protocol: each group includes five decks, and participants typically complete the ranking task within about 3 minutes. Under such constraints, users are unlikely to verify factual correctness against the background materials and instead rely on readily perceivable cues such as structure, layout, and overall coherence.

Therefore, the ablation results should be interpreted as evidence that human preference signals in fast ranking settings are driven more by easily assessable attributes, rather than as evidence that material-dependent criteria are unnecessary. In contrast, these criteria remain crucial for objective evaluation, since they explicitly reward source grounding and penalize hallucinations and factual errors that may be missed in quick, perception-driven judgments.

\section{Limitations}

Although \Benchmark{} has made encouraging progress in evaluating automated slide generation, the current framework still has several limitations. 

First, we focus on static slide content and do not model temporal presentation factors such as animations or transition pacing. Future work could extend the benchmark to dynamic elements, including animation quality, presentation rhythm, and cross-slide narrative flow, potentially enabled by advances in video understanding.

Second, our tasks mainly reflect general-purpose use cases, with limited coverage of highly specialized domains such as medical or legal reporting.

Finally, although the checklist-based design substantially reduces subjectivity, the evaluation still relies on multimodal LLMs as verifiers, and their capability may limit the reliability of the resulting scores. 
An important direction for future work is to improve the accuracy and robustness of automated verification, so that the benchmark can provide more dependable measurements of real-world, material-grounded slide quality.

\section{Conclusion}

We presented \Benchmark{}, a fine-grained rubric-based benchmark for material-grounded slide deck generation. It includes \BenchmarkSize{} expert-curated instances, each with long-context materials and an instance-specific checklist (54.1 verifiable binary items on average) spanning five quality dimensions. 
\Benchmark{} reframes evaluation as source-grounded, item-level verification with principled aggregation, yielding more reliable and diagnostic signals than holistic judging while posing a greater challenge for slide generation systems.

Experiments across diverse slide generation systems show that \Benchmark{} aligns more closely with human preferences than existing evaluations. Yet material-grounded slide authoring remains challenging, particularly in terms of layout, design, and faithful grounding.
We hope \Benchmark{} will serve as a rigorous testbed and catalyst for developing slide generation systems that are both well-designed and grounded in their background materials.

\bibliographystyle{unsrtnat}
\bibliography{references}

\begin{thebibliography}{28}
\providecommand{\natexlab}[1]{#1}
\providecommand{\url}[1]{\texttt{#1}}
\expandafter\ifx\csname urlstyle\endcsname\relax
  \providecommand{\doi}[1]{doi: #1}\else
  \providecommand{\doi}{doi: \begingroup \urlstyle{rm}\Url}\fi

\bibitem[OpenAI(2023)]{DBLP:journals/corr/abs-2303-08774}
OpenAI.
\newblock {GPT-4} technical report.
\newblock \emph{CoRR}, abs/2303.08774, 2023.
\newblock \doi{10.48550/ARXIV.2303.08774}.
\newblock URL \url{https://doi.org/10.48550/arXiv.2303.08774}.

\bibitem[Zeng et~al.(2024)Zeng, Xu, Wang, Zhang, Yin, Rojas, Feng, Zhao, Lai, Yu, Wang, Sun, Zhang, Cheng, Gui, Tang, Zhang, Li, Zhao, Wu, Zhong, Liu, Huang, Zhang, Zheng, Lu, Duan, Zhang, Cao, Yang, Tam, Zhao, Liu, Xia, Zhang, Gu, Lv, Liu, Liu, Yang, Song, Zhang, An, Xu, Niu, Yang, Li, Bai, Dong, Qi, Wang, Yang, Du, Hou, and Wang]{DBLP:journals/corr/abs-2406-12793}
Aohan Zeng, Bin Xu, Bowen Wang, Chenhui Zhang, Da~Yin, Diego Rojas, Guanyu Feng, Hanlin Zhao, Hanyu Lai, Hao Yu, Hongning Wang, Jiadai Sun, Jiajie Zhang, Jiale Cheng, Jiayi Gui, Jie Tang, Jing Zhang, Juanzi Li, Lei Zhao, Lindong Wu, Lucen Zhong, Mingdao Liu, Minlie Huang, Peng Zhang, Qinkai Zheng, Rui Lu, Shuaiqi Duan, Shudan Zhang, Shulin Cao, Shuxun Yang, Weng~Lam Tam, Wenyi Zhao, Xiao Liu, Xiao Xia, Xiaohan Zhang, Xiaotao Gu, Xin Lv, Xinghan Liu, Xinyi Liu, Xinyue Yang, Xixuan Song, Xunkai Zhang, Yifan An, Yifan Xu, Yilin Niu, Yuantao Yang, Yueyan Li, Yushi Bai, Yuxiao Dong, Zehan Qi, Zhaoyu Wang, Zhen Yang, Zhengxiao Du, Zhenyu Hou, and Zihan Wang.
\newblock Chatglm: {A} family of large language models from {GLM-130B} to {GLM-4} all tools.
\newblock \emph{CoRR}, abs/2406.12793, 2024.
\newblock \doi{10.48550/ARXIV.2406.12793}.
\newblock URL \url{https://doi.org/10.48550/arXiv.2406.12793}.

\bibitem[Team(2024)]{DBLP:journals/corr/abs-2407-21783}
Llama Team.
\newblock The llama 3 herd of models.
\newblock \emph{CoRR}, abs/2407.21783, 2024.
\newblock \doi{10.48550/ARXIV.2407.21783}.
\newblock URL \url{https://doi.org/10.48550/arXiv.2407.21783}.

\bibitem[Yang et~al.(2024)Yang, Yang, Hui, Zheng, Yu, Zhou, Li, Li, Liu, Huang, Dong, Wei, Lin, Tang, Wang, Yang, Tu, Zhang, Ma, Yang, Xu, Zhou, Bai, He, Lin, Dang, Lu, Chen, Yang, Li, Xue, Ni, Zhang, Wang, Peng, Men, Gao, Lin, Wang, Bai, Tan, Zhu, Li, Liu, Ge, Deng, Zhou, Ren, Zhang, Wei, Ren, Liu, Fan, Yao, Zhang, Wan, Chu, Liu, Cui, Zhang, Guo, and Fan]{DBLP:journals/corr/abs-2407-10671}
An~Yang, Baosong Yang, Binyuan Hui, Bo~Zheng, Bowen Yu, Chang Zhou, Chengpeng Li, Chengyuan Li, Dayiheng Liu, Fei Huang, Guanting Dong, Haoran Wei, Huan Lin, Jialong Tang, Jialin Wang, Jian Yang, Jianhong Tu, Jianwei Zhang, Jianxin Ma, Jianxin Yang, Jin Xu, Jingren Zhou, Jinze Bai, Jinzheng He, Junyang Lin, Kai Dang, Keming Lu, Keqin Chen, Kexin Yang, Mei Li, Mingfeng Xue, Na~Ni, Pei Zhang, Peng Wang, Ru~Peng, Rui Men, Ruize Gao, Runji Lin, Shijie Wang, Shuai Bai, Sinan Tan, Tianhang Zhu, Tianhao Li, Tianyu Liu, Wenbin Ge, Xiaodong Deng, Xiaohuan Zhou, Xingzhang Ren, Xinyu Zhang, Xipin Wei, Xuancheng Ren, Xuejing Liu, Yang Fan, Yang Yao, Yichang Zhang, Yu~Wan, Yunfei Chu, Yuqiong Liu, Zeyu Cui, Zhenru Zhang, Zhifang Guo, and Zhihao Fan.
\newblock Qwen2 technical report.
\newblock \emph{CoRR}, abs/2407.10671, 2024.
\newblock \doi{10.48550/ARXIV.2407.10671}.
\newblock URL \url{https://doi.org/10.48550/arXiv.2407.10671}.

\bibitem[Zhang et~al.(2025)Zhang, Ni, Chen, Zhang, Rao, Peng, Lu, Hu, Guo, and Hu]{DBLP:journals/corr/abs-2510-13795}
Yi~Zhang, Bolin Ni, Xin{-}Sheng Chen, Heng{-}Rui Zhang, Yongming Rao, Houwen Peng, Qinglin Lu, Han Hu, Meng{-}Hao Guo, and Shi{-}Min Hu.
\newblock Bee: {A} high-quality corpus and full-stack suite to unlock advanced fully open mllms.
\newblock \emph{CoRR}, abs/2510.13795, 2025.
\newblock \doi{10.48550/ARXIV.2510.13795}.
\newblock URL \url{https://doi.org/10.48550/arXiv.2510.13795}.

\bibitem[Yao et~al.(2023)Yao, Zhao, Yu, Du, Shafran, Narasimhan, and Cao]{DBLP:conf/iclr/YaoZYDSN023}
Shunyu Yao, Jeffrey Zhao, Dian Yu, Nan Du, Izhak Shafran, Karthik~R. Narasimhan, and Yuan Cao.
\newblock React: Synergizing reasoning and acting in language models.
\newblock In \emph{The Eleventh International Conference on Learning Representations, {ICLR} 2023, Kigali, Rwanda, May 1-5, 2023}. OpenReview.net, 2023.
\newblock URL \url{https://openreview.net/forum?id=WE\_vluYUL-X}.

\bibitem[Hong et~al.(2024)Hong, Zhuge, Chen, Zheng, Cheng, Wang, Zhang, Wang, Yau, Lin, Zhou, Ran, Xiao, Wu, and Schmidhuber]{DBLP:conf/iclr/HongZCZCWZWYLZR24}
Sirui Hong, Mingchen Zhuge, Jonathan Chen, Xiawu Zheng, Yuheng Cheng, Jinlin Wang, Ceyao Zhang, Zili Wang, Steven Ka~Shing Yau, Zijuan Lin, Liyang Zhou, Chenyu Ran, Lingfeng Xiao, Chenglin Wu, and J{\"{u}}rgen Schmidhuber.
\newblock Metagpt: Meta programming for {A} multi-agent collaborative framework.
\newblock In \emph{The Twelfth International Conference on Learning Representations, {ICLR} 2024, Vienna, Austria, May 7-11, 2024}. OpenReview.net, 2024.
\newblock URL \url{https://openreview.net/forum?id=VtmBAGCN7o}.

\bibitem[Wang et~al.(2025{\natexlab{a}})Wang, Li, Song, Xu, Tang, Zhuge, Pan, Song, Li, Singh, Tran, Li, Ma, Zheng, Qian, Shao, Muennighoff, Zhang, Hui, Lin, and et~al.]{DBLP:conf/iclr/0001LSXTZPSLSTL25}
Xingyao Wang, Boxuan Li, Yufan Song, Frank~F. Xu, Xiangru Tang, Mingchen Zhuge, Jiayi Pan, Yueqi Song, Bowen Li, Jaskirat Singh, Hoang~H. Tran, Fuqiang Li, Ren Ma, Mingzhang Zheng, Bill Qian, Yanjun Shao, Niklas Muennighoff, Yizhe Zhang, Binyuan Hui, Junyang Lin, and et~al.
\newblock Openhands: An open platform for {AI} software developers as generalist agents.
\newblock In \emph{The Thirteenth International Conference on Learning Representations, {ICLR} 2025, Singapore, April 24-28, 2025}. OpenReview.net, 2025{\natexlab{a}}.
\newblock URL \url{https://openreview.net/forum?id=OJd3ayDDoF}.

\bibitem[Huang et~al.(2025)Huang, Liu, Hu, Zhang, and Liu]{huang2025pptbench}
Zheng Huang, Xukai Liu, Tianyu Hu, Kai Zhang, and Ye~Liu.
\newblock Pptbench: Towards holistic evaluation of large language models for powerpoint layout and design understanding.
\newblock \emph{arXiv preprint arXiv:2512.02624}, 2025.

\bibitem[Ofengenden et~al.(2025)Ofengenden, Man, Pang, and Wang]{ofengenden2025pptarena}
Michael Ofengenden, Yunze Man, Ziqi Pang, and Yu-Xiong Wang.
\newblock Pptarena: A benchmark for agentic powerpoint editing.
\newblock \emph{arXiv preprint arXiv:2512.03042}, 2025.

\bibitem[Ge et~al.(2025)Ge, Wang, Zhou, Peng, Subramanian, Tan, Sap, Suhr, Fried, Neubig, and Darrell]{DBLP:conf/cvpr/GeWZPSTSSFND25}
Jiaxin Ge, Zora~Zhiruo Wang, Xuhui Zhou, Yi{-}Hao Peng, Sanjay Subramanian, Qinyue Tan, Maarten Sap, Alane Suhr, Daniel Fried, Graham Neubig, and Trevor Darrell.
\newblock Autopresent: Designing structured visuals from scratch.
\newblock In \emph{{IEEE/CVF} Conference on Computer Vision and Pattern Recognition, {CVPR} 2025, Nashville, TN, USA, June 11-15, 2025}, pages 2902--2911. Computer Vision Foundation / {IEEE}, 2025.
\newblock \doi{10.1109/CVPR52734.2025.00276}.
\newblock URL \url{https://openaccess.thecvf.com/content/CVPR2025/html/Ge\_AutoPresent\_Designing\_Structured\_Visuals\_from\_Scratch\_CVPR\_2025\_paper.html}.

\bibitem[Zheng et~al.(2025)Zheng, Guan, Kong, Zhang, Zheng, Zhou, Lin, Lu, Han, and Sun]{DBLP:conf/emnlp/ZhengGKZZZLLHS25}
Hao Zheng, Xinyan Guan, Hao Kong, Wenkai Zhang, Jia Zheng, Weixiang Zhou, Hongyu Lin, Yaojie Lu, Xianpei Han, and Le~Sun.
\newblock Pptagent: Generating and evaluating presentations beyond text-to-slides.
\newblock In Christos Christodoulopoulos, Tanmoy Chakraborty, Carolyn Rose, and Violet Peng, editors, \emph{Proceedings of the 2025 Conference on Empirical Methods in Natural Language Processing, {EMNLP} 2025, Suzhou, China, November 4-9, 2025}, pages 14402--14418. Association for Computational Linguistics, 2025.
\newblock \doi{10.18653/V1/2025.EMNLP-MAIN.728}.
\newblock URL \url{https://doi.org/10.18653/v1/2025.emnlp-main.728}.

\bibitem[Zheng et~al.(2023)Zheng, Chiang, Sheng, Zhuang, Wu, Zhuang, Lin, Li, Li, Xing, Zhang, Gonzalez, and Stoica]{DBLP:conf/nips/ZhengC00WZL0LXZ23}
Lianmin Zheng, Wei{-}Lin Chiang, Ying Sheng, Siyuan Zhuang, Zhanghao Wu, Yonghao Zhuang, Zi~Lin, Zhuohan Li, Dacheng Li, Eric~P. Xing, Hao Zhang, Joseph~E. Gonzalez, and Ion Stoica.
\newblock Judging llm-as-a-judge with mt-bench and chatbot arena.
\newblock In Alice Oh, Tristan Naumann, Amir Globerson, Kate Saenko, Moritz Hardt, and Sergey Levine, editors, \emph{Advances in Neural Information Processing Systems 36: Annual Conference on Neural Information Processing Systems 2023, NeurIPS 2023, New Orleans, LA, USA, December 10 - 16, 2023}, 2023.
\newblock URL \url{http://papers.nips.cc/paper\_files/paper/2023/hash/91f18a1287b398d378ef22505bf41832-Abstract-Datasets\_and\_Benchmarks.html}.

\bibitem[Chen et~al.(2024)Chen, Chen, Zhang, Wang, Liu, Zhou, Zhang, Wan, Zhou, and Sun]{DBLP:conf/icml/ChenCZWLZZ00024}
Dongping Chen, Ruoxi Chen, Shilin Zhang, Yaochen Wang, Yinuo Liu, Huichi Zhou, Qihui Zhang, Yao Wan, Pan Zhou, and Lichao Sun.
\newblock Mllm-as-a-judge: Assessing multimodal llm-as-a-judge with vision-language benchmark.
\newblock In \emph{Forty-first International Conference on Machine Learning, {ICML} 2024, Vienna, Austria, July 21-27, 2024}. OpenReview.net, 2024.
\newblock URL \url{https://openreview.net/forum?id=dbFEFHAD79}.

\bibitem[Yang et~al.(2026)Yang, Li, Ren, Lu, Wang, Huang, Zong, Zhan, and Li]{yang2026slidesgen}
Yunqiao Yang, Wenbo Li, Houxing Ren, Zimu Lu, Ke~Wang, Zhiyuan Huang, Zhuofan Zong, Mingjie Zhan, and Hongsheng Li.
\newblock Slidesgen-bench: Evaluating slides generation via computational and quantitative metrics.
\newblock \emph{arXiv preprint arXiv:2601.09487}, 2026.

\bibitem[{Google}(2025)]{notebooklm}
{Google}.
\newblock Notebooklm.
\newblock \url{https://notebooklm.google}, 2025.
\newblock Accessed: 2026-02-08.

\bibitem[Li et~al.(2024{\natexlab{a}})Li, Dong, Chen, Su, Zhou, Ai, Ye, and Liu]{DBLP:journals/corr/abs-2412-05579}
Haitao Li, Qian Dong, Junjie Chen, Huixue Su, Yujia Zhou, Qingyao Ai, Ziyi Ye, and Yiqun Liu.
\newblock Llms-as-judges: {A} comprehensive survey on llm-based evaluation methods.
\newblock \emph{CoRR}, abs/2412.05579, 2024{\natexlab{a}}.
\newblock \doi{10.48550/ARXIV.2412.05579}.
\newblock URL \url{https://doi.org/10.48550/arXiv.2412.05579}.

\bibitem[Li et~al.(2024{\natexlab{b}})Li, Sun, Yuan, Fan, Zhao, and Liu]{DBLP:conf/iclr/LiSYF0024}
Junlong Li, Shichao Sun, Weizhe Yuan, Run{-}Ze Fan, Hai Zhao, and Pengfei Liu.
\newblock Generative judge for evaluating alignment.
\newblock In \emph{The Twelfth International Conference on Learning Representations, {ICLR} 2024, Vienna, Austria, May 7-11, 2024}. OpenReview.net, 2024{\natexlab{b}}.
\newblock URL \url{https://openreview.net/forum?id=gtkFw6sZGS}.

\bibitem[Kim et~al.(2024)Kim, Suk, Longpre, Lin, Shin, Welleck, Neubig, Lee, Lee, and Seo]{DBLP:conf/emnlp/KimSLLSWNL0S24}
Seungone Kim, Juyoung Suk, Shayne Longpre, Bill~Yuchen Lin, Jamin Shin, Sean Welleck, Graham Neubig, Moontae Lee, Kyungjae Lee, and Minjoon Seo.
\newblock Prometheus 2: An open source language model specialized in evaluating other language models.
\newblock In Yaser Al{-}Onaizan, Mohit Bansal, and Yun{-}Nung Chen, editors, \emph{Proceedings of the 2024 Conference on Empirical Methods in Natural Language Processing, {EMNLP} 2024, Miami, FL, USA, November 12-16, 2024}, pages 4334--4353. Association for Computational Linguistics, 2024.
\newblock \doi{10.18653/V1/2024.EMNLP-MAIN.248}.
\newblock URL \url{https://doi.org/10.18653/v1/2024.emnlp-main.248}.

\bibitem[Wang et~al.(2025{\natexlab{b}})Wang, Yin, Zhao, Tian, Qiao, Wang, Dai, and Luo]{DBLP:journals/corr/abs-2509-14232}
Zhaokai Wang, Penghao Yin, Xiangyu Zhao, Changyao Tian, Yu~Qiao, Wenhai Wang, Jifeng Dai, and Gen Luo.
\newblock Genexam: {A} multidisciplinary text-to-image exam.
\newblock \emph{CoRR}, abs/2509.14232, 2025{\natexlab{b}}.
\newblock \doi{10.48550/ARXIV.2509.14232}.
\newblock URL \url{https://doi.org/10.48550/arXiv.2509.14232}.

\bibitem[Luo et~al.(2025)Luo, Yuan, Chen, Cai, Yue, Yang, Daha, Li, and Lian]{luo2025mmmg}
Yuxuan Luo, Yuhui Yuan, Junwen Chen, Haonan Cai, Ziyi Yue, Yuwei Yang, Fatima~Zohra Daha, Ji~Li, and Zhouhui Lian.
\newblock {MMMG}: A massive, multidisciplinary, multi-tier generation benchmark for text-to-image reasoning.
\newblock In \emph{The Thirty-ninth Annual Conference on Neural Information Processing Systems Datasets and Benchmarks Track}, 2025.
\newblock URL \url{https://openreview.net/forum?id=kzXM6GtU3g}.

\bibitem[Chang et~al.(2025)Chang, Fang, Xing, Wu, Cheng, Wang, Zeng, YU, and Chen]{chang2025oneigbench}
Jingjing Chang, Yixiao Fang, Peng Xing, Shuhan Wu, Wei Cheng, Rui Wang, Xianfang Zeng, Gang YU, and Hai-Bao Chen.
\newblock One{IG}-bench: Omni-dimensional nuanced evaluation for image generation.
\newblock In \emph{The Thirty-ninth Annual Conference on Neural Information Processing Systems Datasets and Benchmarks Track}, 2025.
\newblock URL \url{https://openreview.net/forum?id=S9TQM1Uhpl}.

\bibitem[{Manus}(2025)]{manus}
{Manus}.
\newblock Manus.
\newblock \url{https://manus.im/app}, 2025.
\newblock Accessed: 2026-02-08.

\bibitem[{Gamma}(2025)]{gamma}
{Gamma}.
\newblock Gamma.
\newblock \url{https://gamma.app}, 2025.
\newblock Accessed: 2026-02-08.

\bibitem[{ByteDance}(2025)]{doubao}
{ByteDance}.
\newblock Doubao.
\newblock \url{https://www.doubao.com/chat}, 2025.
\newblock Accessed: 2026-02-08.

\bibitem[{Kunlun}(2025)]{tiangong}
{Kunlun}.
\newblock Tiangong.
\newblock \url{https://www.tiangong.cn}, 2025.
\newblock Accessed: 2026-02-08.

\bibitem[{Z.ai}(2025)]{zhipu}
{Z.ai}.
\newblock Chatglm.
\newblock \url{https://chatglm.cn/main/alltoolsdetail}, 2025.
\newblock Accessed: 2026-02-08.

\bibitem[{Alibaba}(2025)]{qwen}
{Alibaba}.
\newblock Qwen.
\newblock \url{https://www.qianwen.com/aippt}, 2025.
\newblock Accessed: 2026-02-08.

\end{thebibliography}

\clearpage
\onecolumn
\appendix
\section{Appendix}
\subsection{Detailed Composition of the Dataset}  

\cref{tab:detailed_composition_of_the_dataset} summarizes the composition of the dataset by category, source, language, and sample count.

\begin{longtable}{@{}lllr@{}} 
    \caption{Detailed composition of the dataset.} \label{tab:detailed_composition_of_the_dataset} \\
    \toprule
    \textbf{Category} & \textbf{Data Source} & \textbf{Language} & \textbf{Count} \\
    \midrule
    \endfirsthead

    \multicolumn{4}{c}%
    {\tablename\ \thetable\ -- \textit{Continued from previous page}} \\
    \toprule
    \textbf{Category} & \textbf{Data Source} & \textbf{Language} & \textbf{Count} \\
    \midrule
    \endhead

    \midrule
    \multicolumn{4}{r}{\textit{Continued on next page}} \\
    \endfoot
    \bottomrule
    \endlastfoot

    \multirow{14}{*}{Academia} 
        & ICLR 2025 papers & English & 5 \\
        & ICLR 2024 papers & English & 5 \\
        & ICML 2025 papers & English & 5 \\
        & ICML 2024 papers & English & 5 \\
        & CVPR 2025 papers & English & 10 \\
        & CVPR 2024 papers & English & 5 \\
        & CVPR 2023 papers & English & 5 \\
        & NeurIPS 2024 papers & English & 1 \\
        & USENIX Security 2024 papers & English & 15 \\
        & OSDI 2025 papers & English & 5 \\
        & NSDI 2025 papers & English & 5 \\
        & FAST 2025 papers & English & 5 \\
        & NBER conference papers & English & 10 \\
        & F1000Research papers & English & 10 \\
    \midrule
    \multirow{5}{*}{Advertising}
        & Apple iPhone webpages & English & 3 \\
        & Apple iPad webpages & English & 3 \\
        & Apple Mac webpages & English & 2 \\
        & Lenovo laptop webpages & English & 3 \\
        & BMW car brochures & English & 5 \\
    \midrule
    \multirow{5}{*}{Education}
        & \textit{Computer Systems: A Programmer's Perspective} textbook & English & 20 \\
        & \textit{Computer Science: An Overview} textbook & English & 9 \\
        & Tsinghua University \textit{Data Structure} textbook & Chinese & 11 \\
        & MIT OpenCourseWare course 9.13: \textit{The Human Brain} & English & 10 \\
        & MIT OpenCourseWare course 11.437: \textit{Financing Economic Development} & English & 10 \\
    \midrule
    \multirow{6}{*}{Economics}
        & Microsoft earnings releases & English & 10 \\
        & Alphabet earnings releases & English & 7 \\
        & Tesla quarterly update letters & English & 7 \\
        & JPMorgan Chase earnings releases & English & 7 \\
        & OECD \textit{Economic Outlook} reports & English & 5 \\
        & World Bank \textit{Global Economic Prospects} reports & English & 5 \\
    \midrule
    \multirow{3}{*}{Talk}
        & Middle school English class presentations  & English & 12 \\
        & TED talks in Chinese  & Chinese & 8 \\
        & U.S. presidential speeches  & English & 10 \\
\end{longtable}

\subsection{Instruction Example} \label{sec:instruction_example}

\cref{lst:instruction_example} shows an example of our instructions for slide deck generation. The instruction is manually created based on the corresponding materials collected by experts.

\begin{lstlisting}[language=Markdown, caption={Instruction example for slide deck generation.}, label={lst:instruction_example}]
You are an expert lecturer. Your task is to create a **complete set of lecture slides** intended for **in-class teaching at the undergraduate level**. The slides must faithfully present and explain the content of the provided university-level textbook chapters.

## 1. Structure Requirements
The slide deck MUST have **21-35 slides**.
The slide deck must contain the following **ordered sections**, with specific coverage requirements for each. The number of slides in each section may be determined as appropriate.

For any claims or conclusions presented in the slides, it is not sufficient to state the result alone; an appropriate explanation of the underlying principles or reasons must also be provided.

1. **Title Slide**
* **Course/Book Title:** "Computer Science: An Overview (11th Edition)"
* **Chapter Focus:** Chapter 1: Data Storage
* **Context:** Introduction to the fundamental representation of information in computing systems.

2. **Agenda / Outline**
* **Chapter Theme:** The Transition from Analog Information to Digital Representation.
* **Core Topics:** Bits & Logic, Storage Hardware (Memory/Disks), Data Representation (Text/Media), Binary Arithmetic, and Data Integrity (Compression/Errors).
* **Goal:** To understand how diverse information (text, audio, video, numbers) is homogenized into bit patterns and stored reliably.

3. **Theoretical Foundations: Bits and Boolean Logic (Section 1.1)**
* **The Bit:** Define the bit (binary digit) as a symbol whose meaning depends on application (numeric vs. status).
* **Boolean Operations:** Present the mathematical definitions of AND, OR, XOR, and NOT.
* **Gate Abstraction:** Illustrate how boolean operations map to physical hardware gates.
* **Flip-Flops:** Explain the abstraction of a flip-flop circuit as the fundamental unit of storage (setting 0 or 1).
* **Hexadecimal Notation:** Define the mapping between 4-bit patterns and Hex symbols ($0-F$) to simplify stream representation.

4. **Hardware Reality: Main Memory vs. Mass Storage (Sections 1.2 & 1.3)**
* **Main Memory (RAM):**
* Define memory organization: Cells (typically 8 bits/1 byte) and Addresses.
 * Explain the ordering of bits: Most Significant Bit (High-order) vs. Least Significant Bit (Low-order).
* Discuss RAM volatility and the distinction between DRAM and SDRAM.
* **Measurement:** Clarify the "Power of 2" (1024) vs. "Power of 10" (1000) terminology for KB, MB, GB.
* **Mass Storage:**
* **Magnetic Systems:** Explain disks, tracks, sectors, cylinders, and zoned-bit recording. Define performance metrics: Seek time, Rotation delay (latency), Access time, Transfer rate.
* **Optical Systems:** Contrast CD/DVD spiral tracks with magnetic concentric tracks. Explain the transition from pits/lands to laser reflection.
* **Flash Drives:** Explain the shift to non-volatile silicon dioxide chambers (no moving parts) and the concept of wear (flash memory degradation).
* **Buffering:** Explain the mismatch between logical records (user view) and physical records (disk view), and the role of buffers in bridging this gap.

5. **Data Representation: Text and Multimedia (Section 1.4)**
* **Text Encoding:**
* **ASCII:** 7-bit standard + 0 padding.
* **Unicode:** 16-bit standard to support international languages (65,536 patterns).
* **Images:**
* **Bit Maps:** Pixel representation, RGB encoding (3 bytes/pixel), and Luminance/Chrominance encoding.
* **Vector/Geometric:** Representing images as geometric structures (scalable) vs. fixed pixels (aliasing/grainy).
* **Audio:**
* **Sampling:** Explain the digitization of a continuous wave via sampling rates (e.g., 44,100 samples/sec).
* **MIDI:** Contrast sampled sound with MIDI (encoding the *instruction* to play a note, not the sound itself).

6. **The Binary System (Section 1.5)**
* **Binary Notation:** Explain the positional quantity system (powers of 2: ...8, 4, 2, 1).
* **Algorithm:** Present the division-by-two algorithm for converting positive integers to binary strings.
* **Binary Addition:** Demonstrate binary addition rules ($1+1=10$, carrying the 1).
* **Fractions:** Explain the radix point for fractional values (powers of $2^{-1}, 2^{-2}$, etc.).

7. **Numeric Storage: Integers (Section 1.6)**
* **Two's Complement Notation:**
* Define the sign bit (0=positive, 1=negative).
* **The Algorithm:** Explain the "copy bits right-to-left until first 1, then complement the rest" method for negation.
* **Addition/Subtraction:** Show that subtraction is simply adding a negated value.
* **Overflow:** Define overflow in fixed-width storage (e.g., adding two positives yields a negative).
* **Excess Notation:** Define Excess systems (e.g., Excess 8) where the zero value is shifted.

8. **Numeric Storage: Fractions & Floating Point (Section 1.7)**
* **Floating-Point Components:** Define the three fields: Sign bit, Exponent field (often Excess notation), and Mantissa field.
* **Normalization:** Explain the rule of starting the mantissa with the leftmost 1.
* **Truncation Errors:**
* Illustrate the inability to represent certain values (e.g., $2 \frac{5}{8}$ in a specific small-bit system) due to mantissa size limits.
* Discuss the order of addition problem (adding small numbers to large numbers) leading to precision loss.

9. **Data Compression (Section 1.8)**
* **Lossless vs. Lossy:** Define the distinction.
* **Run-Length Encoding:** Explanation of compressing repetitive sequences.
* **Frequency-Dependent (Huffman):** Using short codes for frequent items (e.g., 'e', 't') and long codes for rare items ('z', 'x').
* **Dictionary Encoding (LZW):** Explain the adaptive dictionary concept using the "xyx xyx" example.
* **Media Specifics:**
* **Images:** GIF (Palette limitations), JPEG (Baseline standard, exploiting eye sensitivity to brightness vs. color).
* **Audio/Video:** MPEG/MP3 (Temporal and Frequency masking).

10. **Data Integrity: Communication Errors (Section 1.9)**
* **Parity Bits:**
* **Odd vs. Even Parity:** How adding one bit allows detection of single-bit errors.
* **Limitations:** Inability to detect even numbers of errors.
* **Error-Correcting Codes:**
* **Hamming Distance:** Define as the number of differing bits between two patterns.
* **Correction Logic:** Explain how a sufficient Hamming distance allows a receiver to determine the intended pattern despite errors.

## 2. Content Constraints
* **Coverage:** You must cover all major concepts, specifically:
* **Definitions:** Every key concept introduced in the text must appear on at least one slide.
* **Formal Models:** Mathematical definitions, rules, invariants, or algebraic properties must be explicitly explained.
* **Examples:** Important worked illustrations from the text must be included and explained step-by-step.
  * **Summaries:** Each major chapter or subsection must end with a concise summary slide.
* For each key diagram, table, or visual example included in the textbook, you must provide a corresponding slide that describes or reconstructs the figure.
* *Note: You may **not omit** material simply because it is technical or detailed.*
* **Faithfulness to Background Materials**: Use only the information in the material. Do not introduce new terminology, external examples, or additional theorems not in the text. You must not fabricate factual content or modify or reinterpret the authors' claims.
* **Accuracy:** All content must be factually accurate, especially quantitative content and facts.
* **Brevity:** Use short, concise phrases, not long paragraphs. Focus on summarizing key facts and events without excessive detail. Bullet points may be used for clarity. If you use bullet points, each slide should have no more than 6 bullet points.
* **Sufficient Depth**: Do not summarize the material in an overly superficial or high-level manner. The slides should preserve essential technical details, key arguments, and substantive insights rather than only presenting vague conclusions.
* **Logical Flow:** The slides should present a clear narrative, starting from early space exploration to recent developments. Ensure there is a clear progression of time and events.
* **Relevance of Information**: You must not add unrelated content.
* **Code & Markup Formatting**: Avoid raw LaTeX or Markdown code unless necessary.
* **Citation & Referencing**: 
* **Every slide** must explicitly indicate which textbook chapter(s) and section(s) its content comes from.
* The attribution must refer to the **most fine-grained section level possible**(e.g., "Section 2.3.2-2.3.5").
* Slides without explicit chapter/section attribution are considered **incorrect**.
* Accurately reference the textbook's diagrams, and examples. If a slide uses data from the textbook, you must clearly indicate the source of the data on that slide (e.g., page xx, Figure xx, Table xx).
* All references (if any) must be placed in the bottom-left corner of the slide.

## 3. Visual & Design

* **Charts and Diagrams:** Use appropriate charts and diagrams where needed to visually present and clarify information, rather than relying only on text (and demos).
  * If the slide includes charts or figures, ensure that all visual elements are clearly annotated (e.g., axes are labeled, units are specified, legends are included where needed, and data points are explained when necessary).
  
* Include **figures or diagrams descriptions** when appropriate, e.g., The chart (from page 4 in the paper) shows proprietary models outperform open-weight ones.

* **Images:** Include relevant images if necessary. Images must be high quality, clearly labeled, and relevant to the content.
* **Legibility:** Use legible fonts and avoid clutter. Text should be large enough to be easily read.

* **Visual Balance:** Balance text and visuals so slides are easy to read when projected.

* **Layout:** Maintain a clean, professional layout with appropriate fonts, colors, and formatting.

* **Style Consistency**: The entire slide deck should follow a unified and coherent visual style.

* **Information Load**: Slides should avoid excessive information per page to preserve readability.

## 4. Text Quality

* All generated text should be clear, with no missing or incorrect characters or words.
* Spelling, grammar, and typography must be accurate and correct throughout the content.

## 5. Technical Fidelity Requirements

The slide deck must include at least 7 slides containing quantitative content, such as mathematical formulas, calculation examples, worked examples, or experimental results. The entire slide deck must not rely solely on high-level natural language explanations.

* Experimental data and constants must exactly match those presented in the textbook.
* Formulas must be identical to those in the textbook, mathematically equivalent, or derivable from textbook formulas.
* All calculations and reasoning must follow the rules and methods described in the textbook and be logically correct.
* Ensure that any figures and tables in your slide deck are consistent with the textbook. Specifically, for every figure and table in the slides:
* If it is directly copied from the textbook, clearly indicate on the slide which figure or table it corresponds to in the textbook (e.g., Figure 1 in the textbook, Table 2 in the textbook).
* If it is newly plotted based on data from the textbook, clearly specify which section of the textbook the data are taken from (e.g., Section 3.1). In addition, provide a clear explanation of the meaning of each legend item in the figure and each row and column in the table.
* Mathematical figures (e.g., function plots) must be logically equivalent to the corresponding textbook content, not merely similar in shape.
* If statistical charts (such scatter plots, line charts or radar charts) are used in the slide deck, ensure that every data point exactly matches the corresponding data point in the original figure from the textbook. Note that the values must be **precisely** the same, not merely that the overall trends align.
* The slides may include data used for conceptual illustration or experimental data. However, You must clearly indicate on the corresponding slide which data are conceptual illustrations or experimental data reported in the textbook.

## 6. Presentation Tone and Audience
* **Tone**:
* Academic, clear, and instructional. 
* Maintain consistent tone and formatting.
* No conversational language, rhetorical questions, emojis, jokes, or storytelling.
* Use precise technical language appropriate for a computer science course.
* **Audience:** Undergraduate students encountering this material for the first time.
* **Goal:** Help students understand core concepts, follow logical derivations, interpret figures/tables, and connect mechanisms to reasoning.
* **Prerequisite:** Avoid assuming prior knowledge beyond standard prerequisite courses.

Your generated slide deck should be able to be used directly in a classroom.

# **Output Expected**
A **complete slide deck** satisfying all constraints above.
\end{lstlisting}

\subsection{Checklist Example} \label{sec:checklist_example}

\cref{lst:material_independent_checklist_example,lst:material_dependent_checklist_example} provide example checklists for \cref{lst:instruction_example}.

\begin{lstlisting}[language=python, caption={A material-independent checklist example.}, label={lst:material_independent_checklist_example}]]
# Material Independent Checklist

material_independent_prefix = r"""You are an expert in evaluating lecture slides.

An AI agent is tasked with creating a complete lecture slide deck based solely on the provided textbook (and other materials, if any). The objective is for the agent to generate a professional, comprehensive, and logically-structured slide deck suitable for **in-class teaching**.

Your task is to **evaluate the slides generated by that AI agent based on the requirement provided below**. The AI-generated slides are provided to you. 

Please indicate whether the generated slides meet the specified requirement by answering "yes" or "no". If no, provide a clear explanation of why it does not meet the requirement. If possible, reference specific slides (e.g., Slide 3, Slide 5) in your explanation.

If the slides fall anywhere between fully meeting and fully failing the requirement (i.e., partially meet it), you MUST classify the answer as "no". Only slides that fully satisfy the requirement with no exceptions may receive "yes".

Your answer must include a `\boxed{...}`, where `...` is "yes" or "no". Aside from this requirement, there are no restrictions on the response format.


Below is the requirement. 

---

"""
## Material Independent Checklist Items
## **1. Content and Structure**
_min_pages, _max_pages = 21, 35
material_independent_checklist_1 = [
# 1.1
# """
# **Is the number of slides between 21-35?**
#   If **no**, explain whether the slides are too few or too many.
# """,
partial(check_slide_count, min_count=_min_pages, max_count=_max_pages),
# 1.2
"""
**Clarity of Key Points**
* Does the slide deck maintain a clear and focused central theme throughout? 
  If **no**, explain where the clarity is lacking.
""",
# 1.3
"""
**Logical Flow**

* Does the slide deck follow a logical progression from one point to the next?

  If **no**, identify specific slides that break the flow.
""",
# 1.4
"""
**Relevance of Information**
* Does each slide contain only the most relevant information, and are the slide titles well aligned with the slide content?

  If **no**, identify slides that contain extraneous or irrelevant details, or whose titles do not accurately reflect their content.
""",
# 1.5
"""
**Avoidance of Placeholder Slides**
* Are there no slides with just an introductory sentence and no real content (e.g., "Introduction to Research")?
  If **no**, mention which slide(s) are too generic or contain placeholders.
""",
# 1.6
"""
**Slide Titles**

* Are the titles clear and do they accurately reflect the content of each slide?

  If **no**, list any titles that are unclear or misleading.
""",
# 1.7
"""
**Conciseness**
* Are the slides concise, with minimal unnecessary wording?
  If **no**, identify slides that are overly verbose.
""",
# 1.8
"""
**Suitability for Lecture Slides**
* Is the slide deck suitable for use as lecture slides in a classroom setting?

  If **no**, explain why the slide deck is not suitable (e.g., inappropriate language style and visual style, inappropriate difficulty level, unclear explanations, poor structure, or lack of alignment with instructional goals).
""",
# 1.9
"""
**Slide-Only Content Compliance**

* Does the slide deck avoid including non-slide content such as scripts, narration, design rationales, or prompts?

  If **no**, specify which slides contain non-slide content and describe the content included.
""",

# 1.10
"""
**Harmful or Biased Content**

* Is the presentation free of harmful or biased content (e.g., images or text involving violence, sexual content, discrimination, illegal material, or anything that may cause psychological discomfort)?

  If **no**, specify which slides contain harmful or biased content.
""",

# 1.11
"""
**Spelling Accuracy**

* Are all words spelled correctly?

  Note: Only evaluate spelling accuracy of words. Do not consider whether the characters or letters themselves are valid or correctly rendered (e.g., existence of characters, garbling), and do not evaluate grammatical correctness.

  If **no**, specify any misspelled words and their location.
""",

# 1.12
"""
**Grammatical Accuracy**

* Are all sentences grammatically correct?

  Note: Only evaluate grammatical correctness. Do not consider whether the characters or letters themselves are valid or correctly rendered (e.g., nonexistent, garbled, or missing characters), and do not evaluate spelling accuracy.

  If **no**, specify the slide number and the exact location where the grammar is incorrect.
""",

# 1.13
"""
**Language Consistency**

* Does the entire slide deck consistently use a single language (e.g., all English or all Chinese) without unintended mixing across slides or within individual slides?

  Note: Occasional use of standard technical terms (e.g., method names, dataset names, or commonly accepted English acronyms) is acceptable, as long as the primary presentation language remains consistent.

  If **no**, specify which slides contain mixed or inconsistent language usage (e.g., English titles with Chinese body text, untranslated labels, or mixed-language bullet points).
""",

]


## **2. Visual Design, Layout, and Graphical Information Representation**

material_independent_checklist_2 = [
# 2.1
"""
**Consistency in Design**

* Is the design consistent across all slides (e.g., font, colors, layout)?

  If **no**, specify which slides deviate from the standard design.
""",

# 2.2
"""
**Balance of Text and Visuals**

* Is there a good balance between text and visuals, avoiding overly text-heavy slides?

  If **no**, indicate which slides are text-heavy or overly reliant on images.
""",

# 2.3
"""
**Decorative Visual Elements**

* Are the decorative visual elements (images, icons, etc.) used in moderation, avoiding an overly busy or cluttered slide design?

  If **no**, specify which slide contains too many decorative elements, making it look overly busy or cluttered.
""",

# 2.4
"""
**Relevance of Visual Elements**

* Are the visual elements (images, icons, etc.) on each slide directly related to the content, contributing meaningfully to the slide's message?

  If **no**, specify which slide includes visual elements (images, icons) that are not closely related to the content of the slide.
""",

# 2.5
"""
**Layout Reasonableness**

* Is the layout reasonable? For example, blank slides, slides that contain only a title without any content, or slides with large areas of empty space (without text or images) are generally inappropriate unless there is a clear justification, such as reserving space for content revealed through animations.

  If **no**, specify which slide has an unreasonable layout and explain why.
""",

# 2.6
"""
**Text and Content Overlap**

* Is all text fully visible and unobstructed, with no overlap with other text or visual elements (images, charts, icons, shapes) that renders the text unreadable or completely obscures it?

  Note: Text with a transparent background image or other visual elements that do not significantly impair readability is not considered a violation. As long as the text remains legible and readable despite the visual elements, this condition is deemed acceptable.

  If **no**, specify the slide number(s) and indicate which text elements are overlapped or occluded.
""",

# 2.7
"""
**Visual Element Overlap**

* Are images, charts, diagrams, and decorative visual elements arranged without overlapping or blocking each other in a way that causes visual clutter or hides important information?

  Note: If a foreground element overlaps a background element, and the background is primarily decorative and does not affect readability, this is considered acceptable. However, if foreground elements overlap each other, causing confusion or visual obstruction, this is considered a violation.
  
  If **no**, specify which slide(s) contain overlapping visual elements and describe the issue.
""",

# 2.8
"""
**Image Quality**

* Are all images, diagrams, and graphs high-quality and legible?

  If **no**, mention specific slides with low-quality visuals.
""",

# 2.9
"""
**Appropriate Visuals**

* Does the slide deck contain appropriate visuals (graphs, tables, diagrams) where necessary?

  If **no**, specify which slides lack proper visuals.
""",

# 2.10
"""
**Visual Appeal**

* Are the slides visually appealing and easy to follow?

  If **no**, mention any slides with excessive text, crowded visuals, or poor design choices.
""",

# 2.11
"""
**Bullet Point Limitation**

* Are no slides overcrowded with more than 6 bullet points (i.e., readable content)?

  If **no**, mention which slide(s) contain excessive information.
""",

# 2.12
"""
**Font Size and Legibility**

* Are the fonts large enough to be easily readable from a distance?

  If **no**, specify any slides where text is too small.
""",

# Graphical Information Representation
# 2.13
"""
**Consistency of Graphical Information Representation**

* Are all graphs, diagrams, flowcharts, charts, and tables presented consistently in terms of style and formatting?

  If the slide deck does not contain graphs, diagrams, flowcharts, charts, or tables, your answer should be "no".

  If **no**, specify any inconsistencies in graphical information representation. 
""",

# 2.14
"""
**Logical Consistency of Graphical Information**

* Are the visuals themselves logically consistent, such that the height of each bar in bar charts or line charts is proportional to the corresponding numerical value, and the angle of each sector in pie charts is proportional to its numerical value? 

  Note: For this criterion, you should assess only the internal logical consistency of the visuals themselves, not whether the data shown matches the values reported in the original material. If the slide deck does not contain graphs, diagrams, flowcharts, charts, and tables, your answer should be "no".

  If **no**, specify which visual elements (e.g. which bar chart / pie chart in which slide) in the charts do not follow the correct proportional relationship. 
""",

# 2.15
"""
**Clarity of Graphical Information**

* Are all charts and figures clearly annotated (i.e., understandable to the audience)?

  For static charts, such as bar charts, line charts, pie charts, and tables, you should check if the axes are labeled, units are specified, legends are included where needed, and data points are explained when necessary.
  If the slide deck does not contain graphs, diagrams, flowcharts, charts, or tables, your answer should be "no".

  If **no**, specify which charts (e.g., bar chart in Slide 4, line plot in Slide 7) lack necessary annotation elements (axis labels, units, legends, captions, etc.).
""",

# 2.16
"""
**Clarity of Text**

* Is all generated text clear, with no missing or incorrect characters or words?

  Note: Only consider whether the characters/letters themselves are valid and correctly rendered (e.g., no nonexistent or garbled characters). Do not consider spelling accuracy or grammatical correctness.

  If **no**, specify the slide number and the exact location where the text is unclear or contains erroneous characters.
""",


# 2.16
"""
**Typographical Accuracy**

* Are all words, labels, axis titles, annotations, and text elements free of typographical errors?

  The slide deck should ensure consistent font, font size and line spacing within the same block of text. All text must use correct and consistent capitalization styles throughout the slides.
  
  Note: Only evaluate typographical and formatting aspects. Do not consider character validity or rendering (e.g., nonexistent or garbled characters), spelling accuracy, or grammatical correctness.

  If **no**, list specific slides and the errors found.
""",

]
\end{lstlisting}

\begin{lstlisting}[language=python, caption={A material-dependent checklist example.}, label={lst:material_dependent_checklist_example}]

material_independent_prefix = r"""You are an expert in evaluating lecture slides.

An AI agent is tasked with creating a complete lecture slide deck based solely on the provided textbook (and other materials, if any). The objective is for the agent to generate a professional, comprehensive, and logically-structured slide deck suitable for **in-class teaching**.

Your task is to **evaluate the slides generated by that AI agent based on the requirement provided below**. The AI-generated slides are provided to you. 

Please indicate whether the generated slides meet the specified requirement by answering "yes" or "no". If no, provide a clear explanation of why it does not meet the requirement. If possible, reference specific slides (e.g., Slide 3, Slide 5) in your explanation.

If the slides fall anywhere between fully meeting and fully failing the requirement (i.e., partially meet it), you MUST classify the answer as "no". Only slides that fully satisfy the requirement with no exceptions may receive "yes".

Your answer must include a `\boxed{...}`, where `...` is "yes" or "no". Aside from this requirement, there are no restrictions on the response format.


Below is the requirement. 

---

"""

# Material Dependent Checklist
material_dependent_checklist_1 = [
"""
**Core Concepts: Bits and Logic**

**Foundational Definitions**
* Does the material define a "bit" as a symbol whose meaning depends entirely on the application at hand?
* Are the four basic Boolean operations explicitly defined: AND, OR, XOR, and NOT?
  Note: Check for the presence of truth tables or logic definitions.
""",
"""
**Hardware Abstraction**
* Is the "Flip-Flop" circuit introduced as the fundamental unit for storing a single bit?
* Does the content explain the abstraction hierarchy: from Gates to Flip-Flops to VLSI (Very Large-Scale Integration) chips?
  If no, specify if the link between logic gates and storage hardware is missing.
""",
"""
**Notation Systems**
* Is Hexadecimal Notation introduced as a shorthand for bit patterns?
* Does the material explain the mapping of 4-bit patterns to single hexadecimal symbols?
""",
"""
**Main Memory and Mass Storage**

**Memory Organization**
* Does the content define "Main Memory" organization, including cells (bytes) and addresses?
* Is the distinction between the "High-order end" (Most Significant Bit) and "Low-order end" (Least Significant Bit) clearly illustrated?
""",
"""
**Mass Storage Technologies**
* Does the material cover the three major classes of mass storage?
  * Magnetic Systems (Disks, Tapes)
  * Optical Systems (CDs, DVDs)
  * Flash Drives (Flash memory, SD cards)
* Are the physical performance metrics defined: Seek time, Rotation delay (latency), Access time, and Transfer rate?
""",
"""
**File Management**
* Is the distinction between "Physical Records" (device blocks) and "Logical Records" (natural data divisions) explained?
* Does the content define "Buffers" and their role in bridging the gap between processor speed and storage transfer rates?
""",
"""
**Data Representation: Text, Images, and Sound**

**Text Encoding Standards**
* Are ASCII and Unicode explicitly contrasted?
* Does the material explain that ASCII uses 7 bits (extended to 8) while Unicode uses 16 bits to support international languages?
""",
"""
**Image Representation**
* Is the difference between "Bit Maps" (Pixel-based) and "Vector/Geometric" representation (Shape-based) explained?
* Does the content detail "RGB encoding" versus "Luminance and Chrominance" encoding?
""",
"""
**Audio Representation**
* Are the two primary methods of audio storage covered?
  * Sampling (recording amplitude at intervals)
  * MIDI (encoding instructions for instruments)
  If no, specify if the distinction between recording sound vs. recording performance instructions is omitted.
""",
"""
**Numeric Representation**

**Integer Storage Systems**
* Does the material introduce the Binary System alongside specific storage notations?
* Are the following notation systems explicitly covered?
  * Two's Complement Notation
  * Excess Notation (e.g., Excess 8)
  If no, specify which integer storage method is missing.
""",
"""
**Fractional Storage**
* Is "Floating-Point Notation" introduced for storing values with fractional parts?
* Does the content identify the three fields within a floating-point representation: Sign bit, Exponent field, and Mantissa field?
""",
"""
**Data Compression and Integrity**

**Compression Techniques**
* Is the distinction between "Lossless" and "Lossy" compression clearly defined?
* Does the material list specific techniques such as Run-length encoding, Frequency-dependent encoding (Huffman), and Dictionary encoding (LZW)?
""",
"""
**Error Management**
* Are methods for handling communication errors included?
* Does the content cover Parity Bits (Odd/Even) for error detection?
* Is the concept of "Hamming Distance" introduced for error correction?
"""
]

material_dependent_checklist_2 = [
"""
**Boolean Logic & Hardware**
* **Gate Logic:** Are the output rules for XOR gates correctly defined (Output is 1 only when inputs are *different*)?
  * *Detail Check:* Ensure XOR is distinguished clearly from OR.
* **Flip-Flop Behavior:** Is the Flip-Flop described as a device that holds a 0 or 1 and changes only in response to an external pulse?
  * *Detail Check:* Verify the description of how setting one input forces an output state that persists after the input is removed.
""",
"""
**Memory Capacity Terminology**
* **Power of 2 vs. Power of 10:** Does the content accurately warn about the ambiguity of prefixes like "Kilo" and "Mega"?
  * *Detail Check:* Verify the statement that in memory contexts, KB = 1024 ($2^{10}$), whereas in communication/speed contexts, it often refers to 1000.
* **Volatility:** Is Main Memory (RAM) correctly described as volatile (requiring refresh circuits for DRAM), contrasting with the non-volatile nature of Mass Storage?
""",
"""
**Mass Storage Dynamics**
* **Disk Geometry:** Are magnetic disks correctly described using "Tracks," "Sectors," and "Cylinders"?
  * *Detail Check:* Check for the mention of "Zoned-bit recording" to explain why outer tracks hold more sectors than inner tracks.
* **Optical Spirals:** Is the physical track structure of CDs/DVDs accurately described as a single spiral (inside-out) rather than concentric circles?
""",
"""
**Two's Complement & Integer Math**
* **Negation Algorithm:** Is the algorithm for finding a negative number in Two's Complement accurately described?
  * *Detail Check:* The text specifies "Copy bits from right to left up to and including the first 1, then complement the remaining bits." Ensure this specific method is cited rather than just "invert and add 1."
* **Overflow logic:** Is "Overflow" correctly defined as a calculation result falling outside the range of representable values?
  * *Detail Check:* Ensure it mentions that adding two positive numbers can produce a negative result (sign bit error) in overflow scenarios.
""",
"""
**Floating-Point Mechanics**
* **Field Interpretation:** In the 8-bit floating-point model used by the text, are the fields defined correctly?
  * *Detail Check:* Sign bit (1 bit), Exponent (3 bits using Excess notation), Mantissa (4 bits).
* **Normalization Rule:** Is the "Normalized Form" rule explained accurately?
  * *Detail Check:* The mantissa must imply a binary point starting with the leftmost 1 (e.g., bit pattern copied starting from the first '1').
* **Truncation Error:** Is the example of "Truncation Error" (Round-off error) mathematically accurate?
  * *Detail Check:* Does it explain why $2 \frac{5}{8}$ cannot be stored accurately in the 8-bit model due to limited mantissa space?
""",
"""
**Compression Algorithms**
* **LZW Dynamic Dictionary:** Is the LZW algorithm correctly described as "Adaptive Dictionary Encoding"?
  * *Detail Check:* Verify the explanation that the dictionary *grows* during the encoding process based on the message content.
* **JPEG vs. GIF:** Are the mechanisms for image compression accurately distinguished?
  * *Detail Check:* GIF uses a palette (256 colors), while JPEG exploits the human eye's sensitivity to brightness over color (averaging chrominance).
""",
"""
**Error Correction Theory**
* **Parity Limitations:** Does the text accurately state that simple Parity Bits cannot detect an even number of errors?
* **Hamming Distance Logic:** Is the relationship between Hamming Distance and error correction capability explained correctly?
  * *Detail Check:* To detect $x$ errors, distance must be $x+1$. To correct $x$ errors, distance must be $2x+1$.
"""
]



material_dependent_prefix = r"""You are an expert in evaluating lecture slides.

An AI agent is tasked with creating a complete lecture slide deck based solely on the provided textbook (and other materials, if any). The objective is for the agent to generate a professional, comprehensive, and logically structured slide deck suitable for **in-class teaching**.

Your task is to **evaluate the slides generated by the AI agent using the checklist provided below**. The AI-generated slides are provided as File 1, and the original textbook (and other materials, if any) used by the AI agent is provided in the subsequent files.

Please indicate whether the generated slides meet the specified requirement by answering "yes" or "no". If no, provide a clear explanation of why it does not meet the requirement. If possible, reference specific slides (e.g., Slide 3, Slide 5) in your explanation.

If the slides fall anywhere between fully meeting and fully failing the requirement (i.e., partially meet it), you MUST classify the answer as "no". Only slides that fully satisfy the requirement with no exceptions may receive "yes".

Your answer must include a `\boxed{...}`, where `...` is "yes" or "no". Aside from this requirement, there are no restrictions on the response format.


Below is the requirement. 

---

"""



# Material Dependent Checklist

## **3. Avoidance of Fabrication and Omission**

material_dependent_checklist_3 = [
f"""
  Is all content on Slide {i} present in the provided background materials and fully consistent with them-i.e., can every statement, number, figure/table reference, chart element, and quoted phrasing be traced to the materials without omission, fabrication, or contradiction?

  * If certain data are explicitly marked as being used only for conceptual illustration, those data are excluded from the scope of evaluation. All other data, unless explicitly marked as conceptual illustrations, must be checked.
  * For scatter plots, line charts, or radar charts, every data point shown on the slide must exactly match the corresponding data point in the original figure from the materials. Note that the values must be precisely the same, not merely that the overall trends align. Otherwise, your answer should be "no".
  * Formulas must be identical to those in the materials, mathematically equivalent, or derivable from materials formulas.
  * All calculations and reasoning must follow the rules and methods described in the materials and be logically correct.
  * All numerical computations must be correct.

  If **no**, specify which data on the slide does not appear in the materials or which numerical values differ from those in the materials (mention the slide's value and the materials' corresponding value).

"""
    for i in range(1, _max_pages+1)
]
\end{lstlisting}

\subsection{Case Study} \label{sec:case_study}

\begin{figure}[htbp]
\begin{tcolorbox}[
    colback=gblue9!5,
    colframe=gblue9!75,
    left=2mm, right=2mm,
    title=\textcolor{white}{\textbf{Case: Passed Checklist Item with Evidence}},
    enhanced,
    sharp corners,
    box align=center,
    halign title=left
]
\begin{tikzpicture}

    \node[inner sep=0pt] (image) at (0,0) {
        \includegraphics[width=12cm,height=12cm,keepaspectratio]{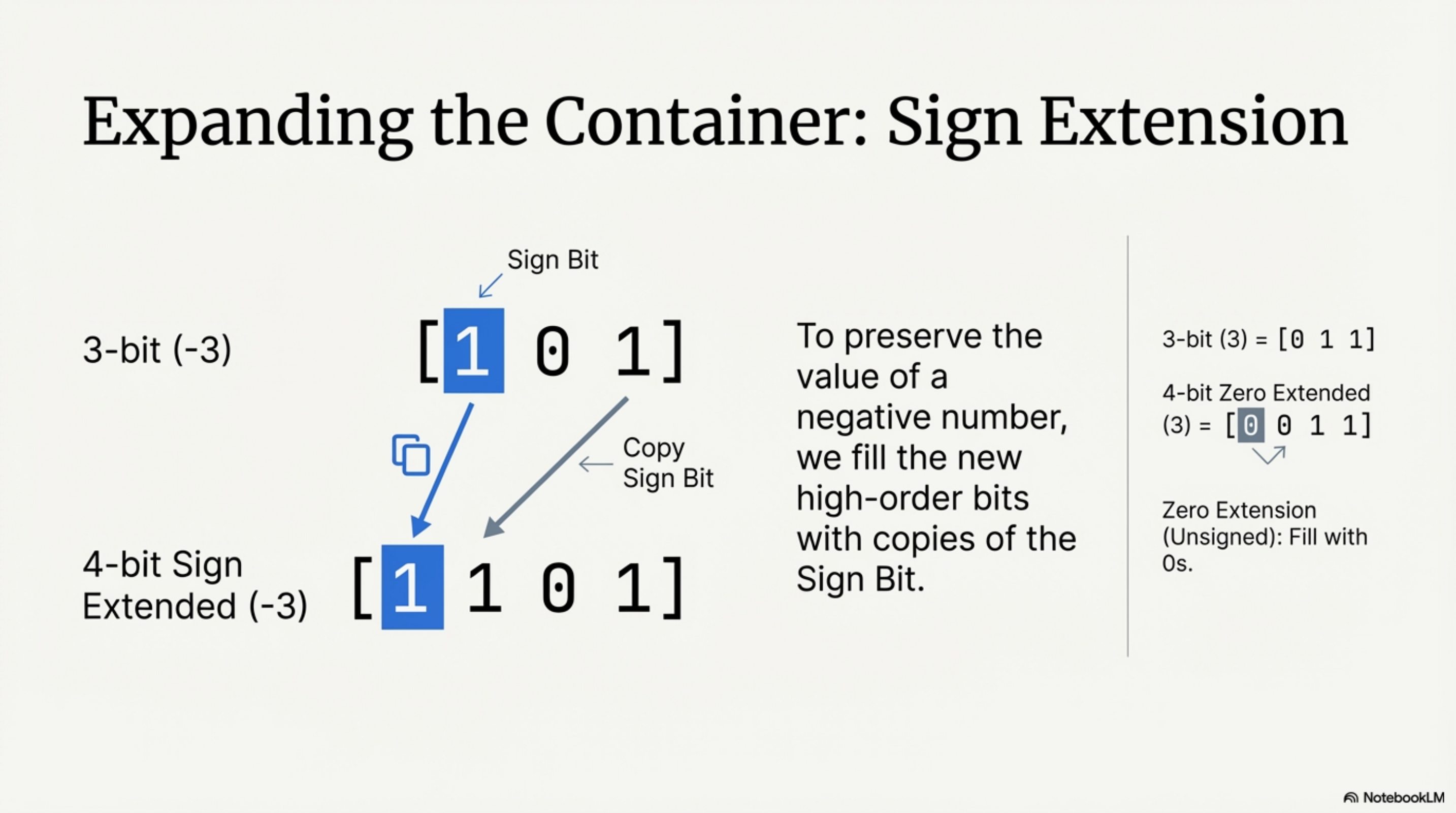}
    };

    \node[below=4mm of image, anchor=north, align=left, text width=\textwidth, font=\bfseries] (item) {Checklist Item:};
    \node[below=1mm of item, anchor=north, align=left, text width=\textwidth] (item_text) {
    **Zero Extension vs. Sign Extension**

    * Are **zero extension** (for unsigned values) and **sign extension** (for two’s complement values) both explained, with their differences clearly highlighted?
    
    Note: You only need to check whether the slides contain the required elements; you do not need to verify their correctness.
    If **no**, specify whether one of them is not covered or whether the distinction between the two is not made explicit.
    };

    \node[below=4mm of item_text, anchor=north, align=left, text width=\textwidth, font=\bfseries] (judgment) {Judgment:};
    \node[below=1mm of judgment, anchor=north, align=left, text width=\textwidth] (judgment_text) {
    Yes
    };

    \node[below=4mm of judgment_text, anchor=north, align=left, text width=\textwidth, font=\bfseries] (explanation) {Explanation: };
    \node[below=1mm of explanation, anchor=north, align=left, text width=\textwidth] (explanation_text) {
        The AI-generated slides meet the requirement for explaining and distinguishing between zero extension and sign extension.

        Specifically:

        * **Sign Extension** is explained on **Slide 9** with the rule: "To preserve
        the value of a negative number, we fill the new high-order bits with copies
        of the Sign Bit." It includes a visual example of a 3-bit number becoming
        a 4-bit number.

        ...
        
        \fbox{yes}
    };

\end{tikzpicture}

\end{tcolorbox}
\caption{The evaluation report provides a detailed explanation for the passed checklist item.}
\label{fig:case_study_1}
\end{figure}

\begin{figure}[htbp]
\begin{tcolorbox}[
    colback=gblue9!5,
    colframe=gblue9!75,
    left=2mm, right=2mm,
    title=\textcolor{white}{\textbf{Case: Missing Content}},
    enhanced,
    sharp corners,
    box align=center,
    halign title=left
]
\begin{tikzpicture}

    \node[inner sep=0pt] (image) at (0,0) {
        \includegraphics[width=12cm,height=12cm,keepaspectratio]{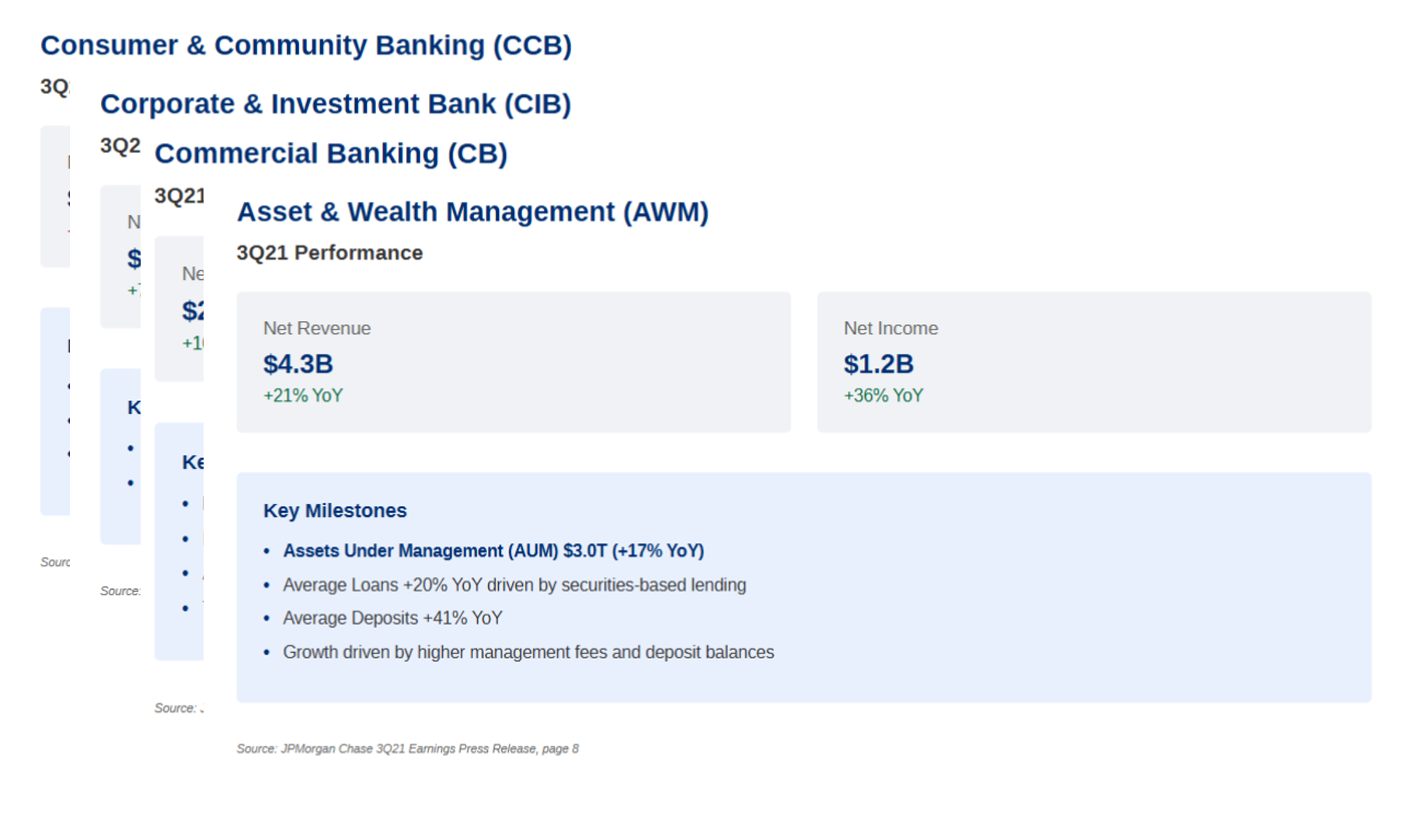}
    };
    \node[below=2mm of image, anchor=north, align=center, text width=12cm, font=\scriptsize] (imgcap) {(Slides 7--10)};
    
    \node[below=6mm of image, anchor=north, align=left, text width=\textwidth, font=\bfseries] (item) {Checklist Item:};
    \node[below=1mm of item, anchor=north, align=left, text width=\textwidth] (item_text) {
    **For each segment, are revenue and net income figures presented?**
    
    Each segment should include:
    
    * Net Revenue
    
    * Net Income
    
    * Year-over-year changes or growth trends
    
      Note: You only need to check whether the slides contain the required content; you do not need to verify their correctness.
      If **no**, specify which quantitative elements are missing for which segment.
    };

    \node[below=4mm of item_text, anchor=north, align=left, text width=\textwidth, font=\bfseries] (judgment) {Judgment:};
    \node[below=1mm of judgment, anchor=north, align=left, text width=\textwidth] (judgment_text) {
    No
    };

    \node[below=4mm of judgment_text, anchor=north, align=left, text width=\textwidth, font=\bfseries] (explanation) {Explanation: };
    \node[below=1mm of explanation, anchor=north, align=left, text width=\textwidth] (explanation_text) {
        The AI-generated slides do not fully meet the requirement because they omit one of the segments present in the original material.

        The source material (File 2) identifies five distinct segments for JPMorgan Chase: **Consumer \& Community Banking (CCB)**, **Corporate \& Investment Bank (CIB)**, **Commercial Banking (CB)**, **Asset \& Wealth Management (AWM)**, and **Corporate** (see File 2, pages 3–5).

        \begin{color}{red}
        While the AI-generated deck includes dedicated slides for four of these segments (Slides 7, 8, 9, and 10), it completely omits the **Corporate** segment. 
        \end{color}

        ...

        \fbox{no}
    };

\end{tikzpicture}

\end{tcolorbox}
\caption{The evaluation report identifies missing content.}
\label{fig:case_study_2}
\end{figure}

\begin{figure}[htbp]
\begin{tcolorbox}[
    colback=gblue9!5,
    colframe=gblue9!75,
    left=2mm, right=2mm,
    title=\textcolor{white}{\textbf{Case: Content Inconsistencies}},
    enhanced,
    sharp corners,
    box align=center,
    halign title=left
]
\begin{tikzpicture}

    \node[anchor=north] (pair) at (0,0) {%
      \begin{minipage}{\textwidth}
        \centering
        \begin{minipage}{0.6\textwidth}
          \centering
          \includegraphics[height=6cm,keepaspectratio]{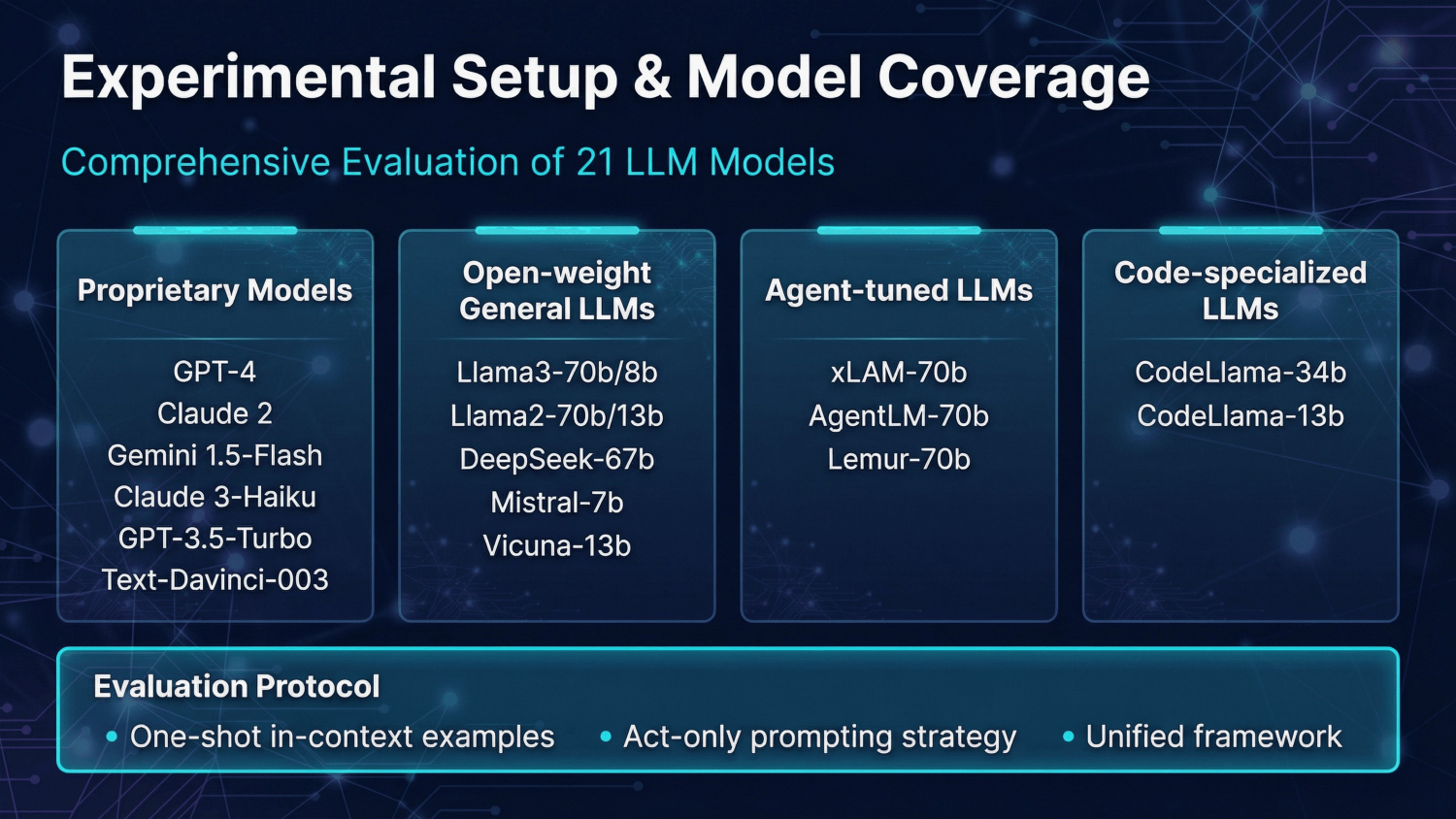}\\[-2pt]
          {\scriptsize (Generated slides)}
        \end{minipage}\hspace{6mm}
        \begin{minipage}{0.25\textwidth}
          \centering
          \includegraphics[height=6cm,keepaspectratio]{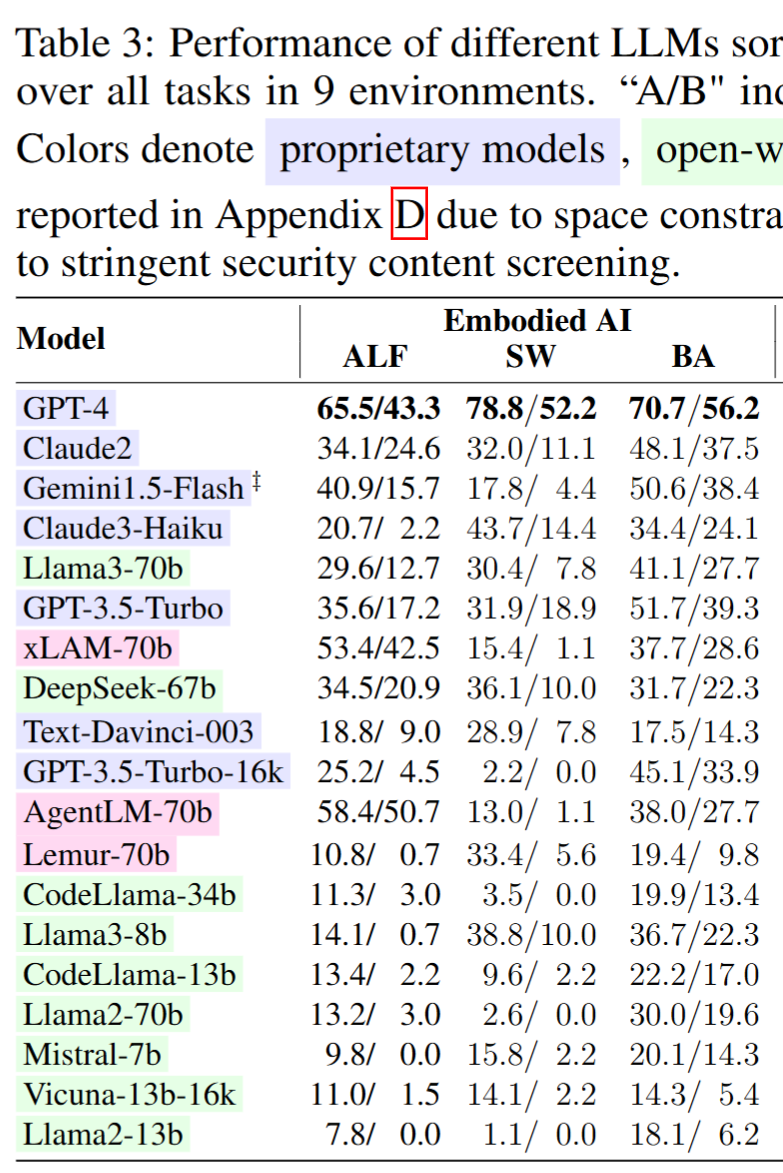}\\[-2pt]
          {\scriptsize (Background material)}
        \end{minipage}
      \end{minipage}
    };

    \node[below=4mm of pair, anchor=north, align=left, text width=\textwidth, font=\bfseries] (item) {Checklist Item:};
    \node[below=1mm of item, anchor=north, align=left, text width=\textwidth] (item_text) {
      Is all content on Slide 8 present in the provided source materials and fully consistent with them, i.e., can every statement, number, figure/table reference, chart element, and quoted phrasing be traced to the materials without omission, fabrication, or contradiction?
    
      * If certain data are explicitly marked as being used only for conceptual illustration, those data are excluded from the scope of evaluation. All other data, unless explicitly marked as conceptual illustrations, must be checked.
      
      * For scatter plots, line charts, or radar charts, every data point shown on the slide must exactly match the corresponding data point in the original figure from the paper. Note that the values must be precisely the same, not merely that the overall trends align. Otherwise, your answer should be "no".
      
      * Formulas must be identical to those in the materials, mathematically equivalent, or derivable from materials formulas.
      
      * All calculations and reasoning must follow the rules and methods described in the materials and be logically correct.
      
      * All numerical computations must be correct.
    
      If **no**, specify which data on the slide does not appear in the paper or which numerical values differ from those in the paper (mention the slide's value and the paper's corresponding value).
    };

    \node[below=4mm of item_text, anchor=north, align=left, text width=\textwidth, font=\bfseries] (judgment) {Judgment:};
    \node[below=1mm of judgment, anchor=north, align=left, text width=\textwidth] (judgment_text) {
    No
    };

    \node[below=4mm of judgment_text, anchor=north, align=left, text width=\textwidth, font=\bfseries] (explanation) {Explanation: };
    \node[below=1mm of explanation, anchor=north, align=left, text width=\textwidth] (explanation_text) {
        The provided slides do not fully meet the requirements. While Slide 8 accurately captures the evaluation protocol (one-shot, act-only, unified framework) mentioned in the paper, it contains several numerical and categorical discrepancies when compared to the source materials:

        \begin{color}{red}
        *   **Fabrication of Model Count:** The subtitle on Slide 8 claims a "Comprehensive Evaluation of **21** LLM Models." However, the source material (Table 3 on page 7) lists a total of **19** models. There is no mention of 21 models in the provided text.
        \end{color}

        ...

        \fbox{no}
    };

\end{tikzpicture}

\end{tcolorbox}
\caption{The evaluation report identifies specific content errors and inconsistencies between the generated slides and the background materials.}
\label{fig:case_study_3}
\end{figure}

\begin{figure}[htbp]
\begin{tcolorbox}[
    colback=gblue9!5,
    colframe=gblue9!75,
    left=2mm, right=2mm,
    title=\textcolor{white}{\textbf{Case: Design Inconsistencies}},
    enhanced,
    sharp corners,
    box align=center,
    halign title=left
]
\begin{tikzpicture}

    \node[anchor=north] (pair) at (0,0) {%
      \begin{minipage}{\textwidth}
        \centering
        \begin{minipage}{0.4\textwidth}
          \centering
          \includegraphics[height=4cm,keepaspectratio]{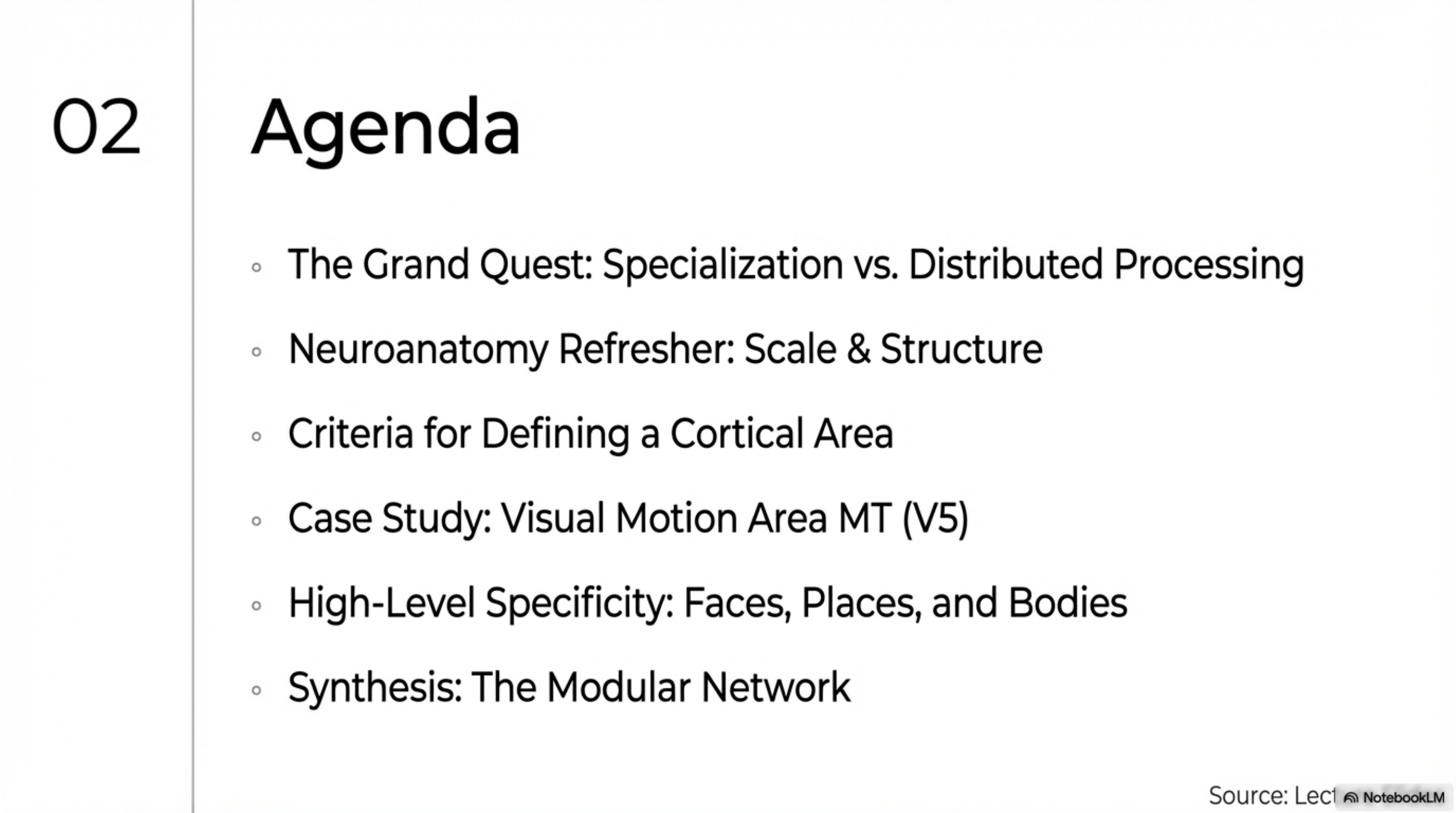}\\[-2pt]
          {\scriptsize (Slide 2)}
        \end{minipage}\hspace{6mm}
        \begin{minipage}{0.4\textwidth}
          \centering
          \includegraphics[height=4cm,keepaspectratio]{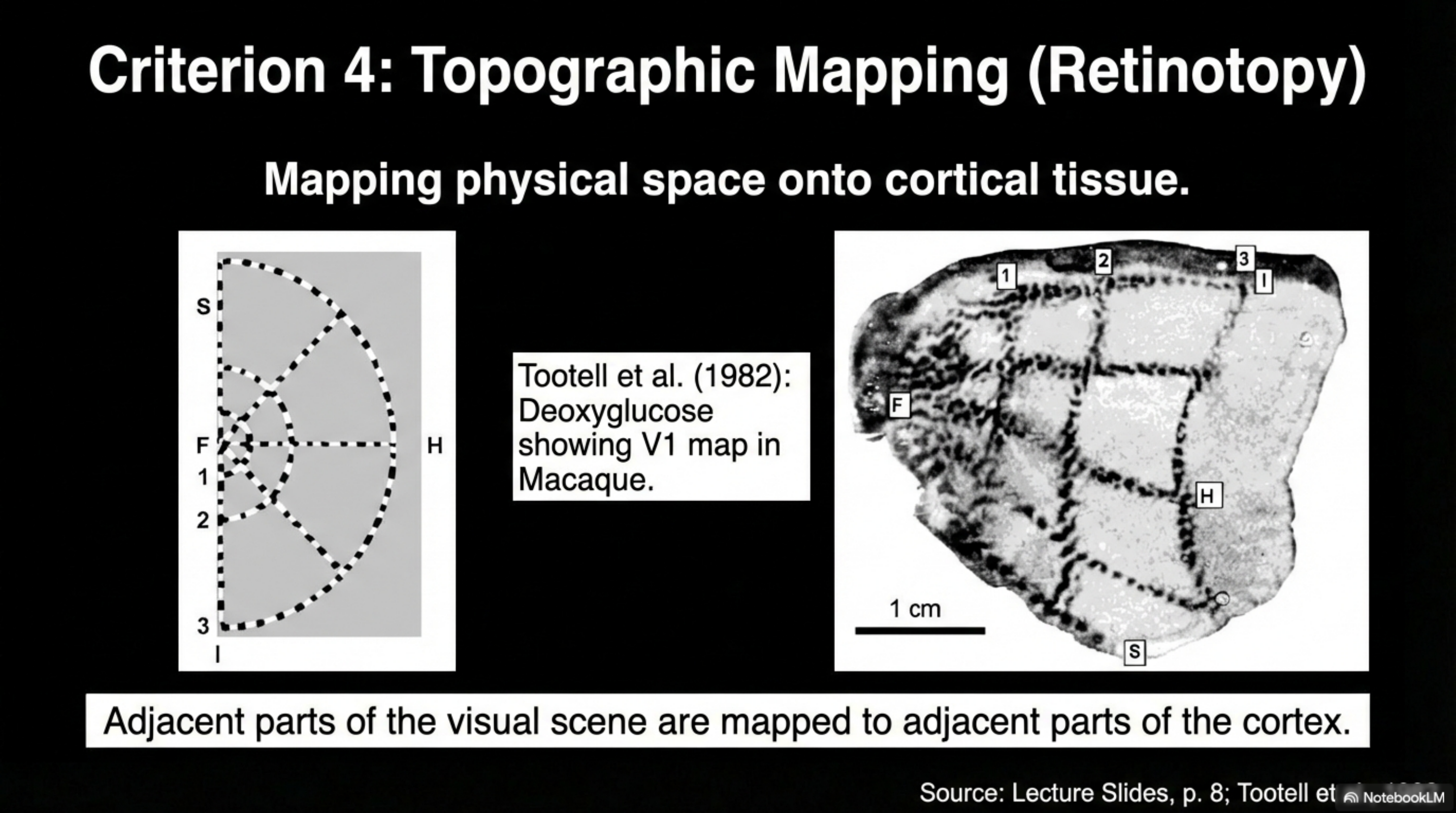}\\[-2pt]
          {\scriptsize (Slide 9)}
        \end{minipage}
      \end{minipage}
    };

    \node[below=4mm of pair, anchor=north, align=left, text width=\textwidth, font=\bfseries] (item) {Checklist Item:};
    \node[below=1mm of item, anchor=north, align=left, text width=\textwidth] (item_text) {
    **Consistency in Design**
    
    * Is the design consistent across all slides (e.g., font, colors, layout)?
    
      If **no**, specify which slides deviate from the standard design.
    };

    \node[below=4mm of item_text, anchor=north, align=left, text width=\textwidth, font=\bfseries] (judgment) {Judgment:};
    \node[below=1mm of judgment, anchor=north, align=left, text width=\textwidth] (judgment_text) {
    No
    };

    \node[below=4mm of judgment_text, anchor=north, align=left, text width=\textwidth, font=\bfseries] (explanation) {Explanation: };
    \node[below=1mm of explanation, anchor=north, align=left, text width=\textwidth] (explanation_text) {
    The design of the slides is inconsistent. Specifically, the background color and overall layout fluctuate throughout the presentation without a clear structural reason, which disrupts the visual flow.

    \begin{color}{red} 
    *   **Background Color Inconsistency:** The deck flips between black and white backgrounds. Slides 1, 6-15, and 18-20 use a black background, while Slides 2-5, 16, and 17 use a white background. 
    \end{color}
    
    *   **Layout Inconsistency:** Slide 2 introduces a specific layout with a large numeral ("02") on the left side, which is never repeated for other sections. 
    
    *   **Font and Title Placement:** On the white-background slides (e.g., Slides 3-5), titles are centered at the top. On the black-background slides (e.g., Slides 11-13), titles vary between centered and left-aligned, and the font sizes for headings are not uniform across the transition from white to black themes.
    
    Because the visual theme (colors, backgrounds, and layout styles) does not remain stable across the entire deck, it does not meet the requirement for consistency.

    \fbox{no}

    };

\end{tikzpicture}

\end{tcolorbox}
\caption{The evaluation report identifies design inconsistencies.}
\label{fig:case_study_4}
\end{figure}

\cref{fig:case_study_1,fig:case_study_2,fig:case_study_3,fig:case_study_4} present representative cases of our fine-grained, rubric-based evaluation results for slide generation.

\end{document}